\documentclass[11pt]{article}

\usepackage[preprint]{acl}

\usepackage{times}
\usepackage{latexsym}

\usepackage[T1]{fontenc}
\usepackage[utf8]{inputenc}

\usepackage{microtype}

\usepackage{inconsolata}

\usepackage{amsmath,amsfonts,bm}

\def\Figref#1{Figure~\ref{#1}}

\def\eqref#1{equation~\ref{#1}}

\def\1{\bm{1}}

\DeclareMathAlphabet{\mathsfit}{\encodingdefault}{\sfdefault}{m}{sl}
\SetMathAlphabet{\mathsfit}{bold}{\encodingdefault}{\sfdefault}{bx}{n}

\usepackage{booktabs}       
\usepackage{amsmath}
\usepackage{amsfonts}       
\usepackage{nicefrac}       
\usepackage{xcolor}         
\usepackage{cleveref}
\usepackage{enumitem}
\usepackage{multirow}
\usepackage{graphicx}
\usepackage{xspace}
\usepackage{adjustbox}
\usepackage{longtable}
\usepackage{array}
\usepackage{colortbl}
\usepackage{subcaption}

\crefname{section}{Section}{Sections}
\crefname{table}{Table}{Tables}
\crefname{figure}{Figure}{Figures}
\crefname{algorithm}{Algorithm}{Algorithms}
\crefname{appendix}{Appendix}{Appendices}
\crefname{equation}{Equation}{Equations}

\let\cite\citep

\usepackage{pifont}
\newcommand{\cmark}{\ding{51}} 
\newcommand{\xmark}{\ding{55}} 

\newcommand{\OurBench}{\textsc{MedFrameQA}\xspace}

\newcommand{\app}{\raise.17ex\hbox{$\scriptstyle\sim$}}
\makeatletter
\DeclareRobustCommand\onedot{\futurelet\@let@token\@onedot}
\def\@onedot{\ifx\@let@token.\else.\null\fi\xspace}
\def\eg{\emph{e.g}\onedot}

\makeatother

\title{MedFrameQA: A Multi-Image Medical VQA Benchmark for Clinical Reasoning}

\author{
{\bfseries
Suhao Yu\textsuperscript{1}\thanks{Equal contribution.} \quad
Haojin Wang\textsuperscript{2}\footnotemark[1] \quad
Juncheng Wu\textsuperscript{3}\footnotemark[1] \quad
Luyang Luo\textsuperscript{4} \quad
Jingshen Wang\textsuperscript{5}
}\\
{\bfseries
Cihang Xie\textsuperscript{3} \quad
Pranav Rajpurkar\textsuperscript{4} \quad
Carl Yang\textsuperscript{6} \quad
Yang Yang\textsuperscript{7} \quad
Kang Wang\textsuperscript{7} \quad
Yannan Yu\textsuperscript{7}
}\\
{\bfseries
Yuyin Zhou\textsuperscript{3}
}\\
\textsuperscript{1}University of Pennsylvania \,
\textsuperscript{2}University of Illinois Urbana-Champaign \,
\textsuperscript{3}UC Santa Cruz \\
\textsuperscript{4}Harvard University\,
\textsuperscript{5}UC Berkeley \,
\textsuperscript{6}Emory University \,
\textsuperscript{7}UC San Francisco
}

\begin{document}

\maketitle

\begin{abstract}
Real-world clinical practice demands multi-image comparative reasoning, yet current medical benchmarks remain limited to single-frame interpretation.
We present \OurBench, the first benchmark explicitly designed to test multi-image medical VQA through educationally-validated diagnostic sequences.
To construct this dataset, we develop a scalable pipeline that leverages narrative transcripts from medical education videos to align visual frames with textual concepts, automatically producing 2,851 high-quality multi-image VQA pairs with explicit, transcript-grounded reasoning chains.
Our evaluation of 11 advanced MLLMs (including reasoning models) exposes severe deficiencies in multi-image synthesis, where accuracies mostly fall below 50\% and exhibit instability across varying image counts.
Error analysis demonstrates that models often treat images as isolated instances, failing to track pathological progression or cross-reference anatomical shifts.
\OurBench provides a rigorous standard for evaluating the next generation of MLLMs in handling complex, temporally grounded medical narratives.
\end{abstract}

\section{Introduction}

\begin{figure*}[t]
    \centering
    \includegraphics[width=\linewidth]{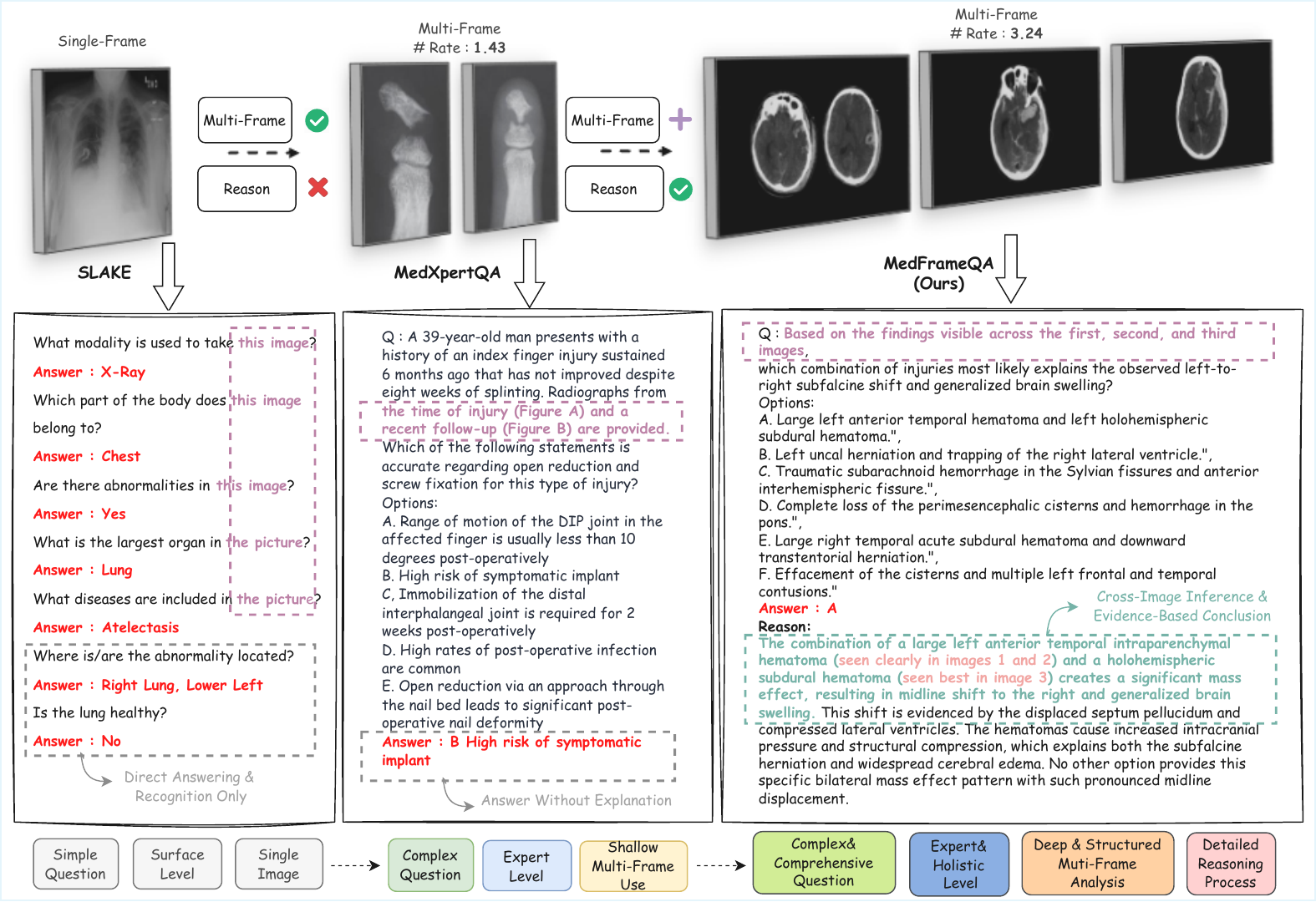}
    \vspace{-1em}
    \caption{\textbf{Comparison of medical VQA benchmarks.}
    \OurBench introduces multi-image, clinically grounded questions that require comprehensive reasoning across all images.
    Unlike prior benchmarks such as SLAKE \citep{DBLP:conf/isbi/LiuZXMYW21} and MedXpertQA \citep{DBLP:journals/corr/abs-2501-18362}, it emphasizes diagnostic complexity, expert-level knowledge, and explicit reasoning chains.
    \textbf{Rate} in the figure means the average number of images per question.
    }
    \vspace{-1em}
    \label{fig:intro}
\end{figure*}

Multimodal Large Language Models (MLLMs) have quickly emerged as a powerful paradigm for enabling advanced AI systems in clinical and medical domains~\citep{xie2025medtrinitym,DBLP:journals/corr/abs-2303-08774,DBLP:conf/nips/LiWZULYNPG23,DBLP:journals/corr/abs-2307-14334,DBLP:journals/corr/abs-2404-18416,huang2025m1,wu2025medreason}. 
In practice, clinicians frequently employ multi-image diagnostic workflows, comparing related scans and synthesizing findings across different views and time points. 
Current evaluation benchmarks, however, focus predominantly on isolated, single-image analysis, \eg,~\citep{lau2018dataset,ImageCLEFVQA-Med2019,ImageCLEF-VQA-Med2021,DBLP:journals/corr/abs-2003-10286,DBLP:conf/isbi/LiuZXMYW21,DBLP:journals/corr/abs-2305-10415,DBLP:conf/cvpr/HuLLSHQL24,DBLP:conf/nips/ChenYWLDLLDHSW024}.
The left panel of \Figref{fig:intro} shows a typical SLAKE~\citep{DBLP:conf/isbi/LiuZXMYW21} example whose answer requires nothing more than basic object recognition in one frame. 
In everyday care, however, clinicians rarely rely on a lone snapshot; they routinely compare multiple images taken from different views, modalities, or time points before making a diagnosis.

Only recently has the multimodal reasoning literature begun to systematically study multi-image VQA. 
A small number of recent benchmarks, such as 
\citet{DBLP:conf/cvpr/YueNZ0LZSJRSWYY24,DBLP:journals/corr/abs-2409-02813,DBLP:journals/corr/abs-2501-18362}, introduce questions that explicitly reference multiple images.
Yet these tasks still fall short of the integrative reasoning required in medicine, as images are often treated as independent clues rather than complementary views of a coherent clinical scenario. 
As illustrated by the MedXpertQA \citep{DBLP:journals/corr/abs-2501-18362} example in the middle panel of \Figref{fig:intro}, the two images lack a clear physiological or causal connection, allowing models to answer correctly without truly synthesizing information across images. 
As a result, performance on such datasets provides limited evidence of a system’s ability to perform clinically grounded cross-image reasoning.

To bridge this gap, we introduce \OurBench, the first benchmark explicitly designed to test multi-image reasoning in medical VQA by leveraging YouTube's rich repository of medical education videos \citep{osman2022youtube, akakpo2024recognizing}. 
Our approach focuses on educational video sequences with temporally and semantically connected visual content that demonstrate diagnostic reasoning within coherent clinical presentations.
Building on this insight and inspired by prior work \citep{DBLP:conf/nips/IkezogwoSGGMAKS23}, we propose a scalable VQA generation pipeline that automatically constructs multi-image VQA questions from keyframes extracted from 3,420 medical videos, covering 9 human body systems and 43 organs. 
We curate medical education videos via combinatorial search, extract and filter keyframes, transcribe and temporally align narrations, and merge clinically related frame–text pairs into coherent multi-frame clips.
Finally, we generate multiple-choice VQA items requiring cross-image clinical reasoning and apply two-stage automated and manual filtering to ensure benchmark quality.

This pipeline yields \OurBench, consisting of 2,851 challenging multi-image VQA questions that require reasoning over temporally coherent sequences of 2–5 frames. 
Each instance pairs a natural-language query with multiple related frames—such as multi-view anatomy, disease progression within educational narratives, or cross-modal comparisons—drawn from continuous medical videos rather than arbitrary image collections (\Figref{fig:intro}, right). 
To support grounded reasoning, we provide gold-standard rationales derived from source video transcripts, explicitly linking each frame to the final answer. 
Beyond the benchmark itself, our work introduces a systematic pipeline for aligning audio narrations with visual frames at scale, offering a new perspective on constructing multimodal reasoning datasets from large video corpora.
We benchmark 11 state-of-the-art MLLMs on \OurBench and find that their accuracies mostly fall below 50\%, with substantial variation across body systems, organs, and modalities, revealing critical gaps between current multimodal model capabilities and the demands of clinically grounded, video-derived multi-image reasoning.



\begin{table*}[t]
    \centering
    \scriptsize
    \label{tab:bench_compare}
    \setlength{\tabcolsep}{4pt}
    \begin{tabular}{lccccccc}
        \toprule
        \textbf{Benchmark} &
        \textbf{\# Images} &
        \textbf{\# Questions} &
        \textbf{\# Rate} &
        \textbf{Multi-Image} &
        \raisebox{-1.8ex}{\shortstack{\textbf{Real World} \\ \textbf{Scenarios}}} &
        \raisebox{-1.8ex}{\shortstack{\textbf{Paired Reasoning} \\ \textbf{Across Multi Img.}}} \\
        \midrule
        VQA-RAD \citep{lau2018dataset} & 315 & 3515 & 0.09 & \xmark & \xmark & \xmark\\
        VQA-Med-2019 \citep{ImageCLEFVQA-Med2019} & 500 & 500 & 1.00 & \xmark & \xmark & \xmark \\
        VQA-Med-2021 \citep{ImageCLEF-VQA-Med2021} & 500 & 500 & 1.00 & \xmark & \xmark & \xmark \\
        PathVQA \citep{DBLP:journals/corr/abs-2003-10286} & 858 & 6,719 & 0.13 & \xmark & \xmark & \xmark \\
        SLAKE-En \citep{DBLP:conf/isbi/LiuZXMYW21} & 96 & 1,061 & 0.09 & \xmark & \xmark & \xmark \\
        PMC-VQA \citep{DBLP:journals/corr/abs-2305-10415} & 29,021 & 33,430 & 0.87 & \xmark & \xmark & \xmark\\
        OmniMedVQA \citep{DBLP:conf/cvpr/HuLLSHQL24} & 118,010 & 127,995 & 0.92 & \xmark & \xmark & \xmark\\
        GMAI-MMBench \citep{DBLP:conf/nips/ChenYWLDLLDHSW024} & 21,180 & 21,281 & 1.00 & \xmark & \xmark & \xmark \\
        MMMU (H\&M) \citep{DBLP:conf/cvpr/YueNZ0LZSJRSWYY24} & 1,994 & 1,752 & 1.14 & \cmark & \xmark & \cmark \\
        MMMU-Pro (H\&M) \citep{DBLP:journals/corr/abs-2409-02813} & 431 & 346 & 1.25 & \cmark & \xmark & \cmark \\
        MedXpertQA MM \citep{DBLP:journals/corr/abs-2501-18362} & 2852 & 2000 & 1.43 & \cmark & \cmark & \xmark \\
        \midrule
        \OurBench & \textbf{9237} & \textbf{2851} & \textbf{3.24} &  \textbf{\cmark} &  \textbf{\cmark} &  \textbf{\cmark} \\
        \bottomrule
    \end{tabular}
    \caption{\textbf{Comparison of \OurBench with Existing Benchmarks.}
    \OurBench supports multi-image reasoning within real-world clinical video scenarios and paired reasoning across frames. The paired reasoning in \OurBench is derived from the transcripts from original video clips.}
    \vspace{-1em}
\end{table*}

\section{Related Work}
\label{Related_work}
\textbf{Reasoning Multimodal Large Language Models\quad}
Recent interest in MLLM reasoning has extended to medical tasks like diagnostics and clinical decision-making \citep{DBLP:journals/corr/abs-2401-06805, xie2024preliminary,chen2025sft,deng2025openvlthinker, AlSaad2024MMLMs, jiang2025hulu}. 
While generalist models like Llava-Med \citep{DBLP:conf/nips/LiWZULYNPG23} and GPT-4V \citep{GPT4V} show promise, they often lack interpretable reasoning. 
To address this, recent works employ strategies such as multi-expert prompting (MedCoT \citep{DBLP:journals/corr/abs-2503-12605}), reinforcement learning for plausible rationales (MedVLM-R1 \citep{DBLP:journals/corr/abs-2502-19634}), and long-context modeling (Med-Gemini \citep{DBLP:journals/corr/abs-2404-18416}, Med-Gemma3 \citep{sellergren2025medgemma}). 
These advances highlight the need for rigorous benchmarks to evaluate medical reasoning capabilities.

\textbf{Multimodal Medical Benchmarks\quad}
Existing benchmarks for evaluating medical MLLMs remain limited in scope, with most focusing on single-image VQA. 
Early datasets such as VQA-RAD \citep{lau2018dataset}, VQA-Med-2019 \citep{ImageCLEFVQA-Med2019}, VQA-Med-2021 \citep{ImageCLEF-VQA-Med2021}, SLAKE \citep{DBLP:conf/isbi/LiuZXMYW21}, and PathVQA \citep{DBLP:journals/corr/abs-2003-10286} primarily target isolated questions within specific medical subdomains. 
More recent benchmarks, including PMC-VQA \citep{DBLP:journals/corr/abs-2305-10415}, OmniMedVQA \citep{DBLP:conf/cvpr/HuLLSHQL24}, and GMAI-MMBench \citep{DBLP:conf/nips/ChenYWLDLLDHSW024}, broaden domain coverage but still largely operate in a single-image setting. 
Although recent efforts such as MMMU (H\&M) \citep{DBLP:conf/cvpr/YueNZ0LZSJRSWYY24}, MMMU-Pro (H\&M) \citep{DBLP:journals/corr/abs-2409-02813}, and MedXpertQA MM \citep{DBLP:journals/corr/abs-2501-18362} incorporate multi-image VQA, they lack clinically grounded cross-image reasoning and gold-standard rationales, limiting their ability to assess genuine multi-image reasoning. 
We provide a detailed comparison with existing benchmarks in \cref{tab:bench_compare}.

\textbf{Video Data For Medical Benchmarking\quad}
Recent work has explored leveraging video data for medical dataset construction, enabled in part by advances in speech recognition models such as Whisper \citep{DBLP:conf/icml/RadfordKXBMS23}. 
Prior efforts have collected large-scale video–text or image–text datasets from medical videos, including Quilt-1M \citep{DBLP:conf/nips/IkezogwoSGGMAKS23}, as well as task-specific benchmarks targeting instructional or procedural content \citep{gupta2023medvidqa,DBLP:conf/nips/HuWYMRXFDJG23,Ghamsarian2024}. 
Despite these advances, video data has seen limited use for benchmarking multimodal large language models (MLLMs) in the medical domain. 
Notably, medical content on YouTube \citep{osman2022youtube,derakhshan2019assessing} naturally provides temporally grounded reasoning chains across frames. 
Motivated by this observation, we leverage YouTube videos and design an automated VQA generation pipeline to construct multi-image questions for evaluating MLLMs under complex, multi-frame clinical scenarios.

\section{MedFrameQA Benchmark}
\label{methods}

\subsection{Video Collection}\label{video_collection}
\begin{figure*}[t]
  \centering
  \includegraphics[width=\linewidth]{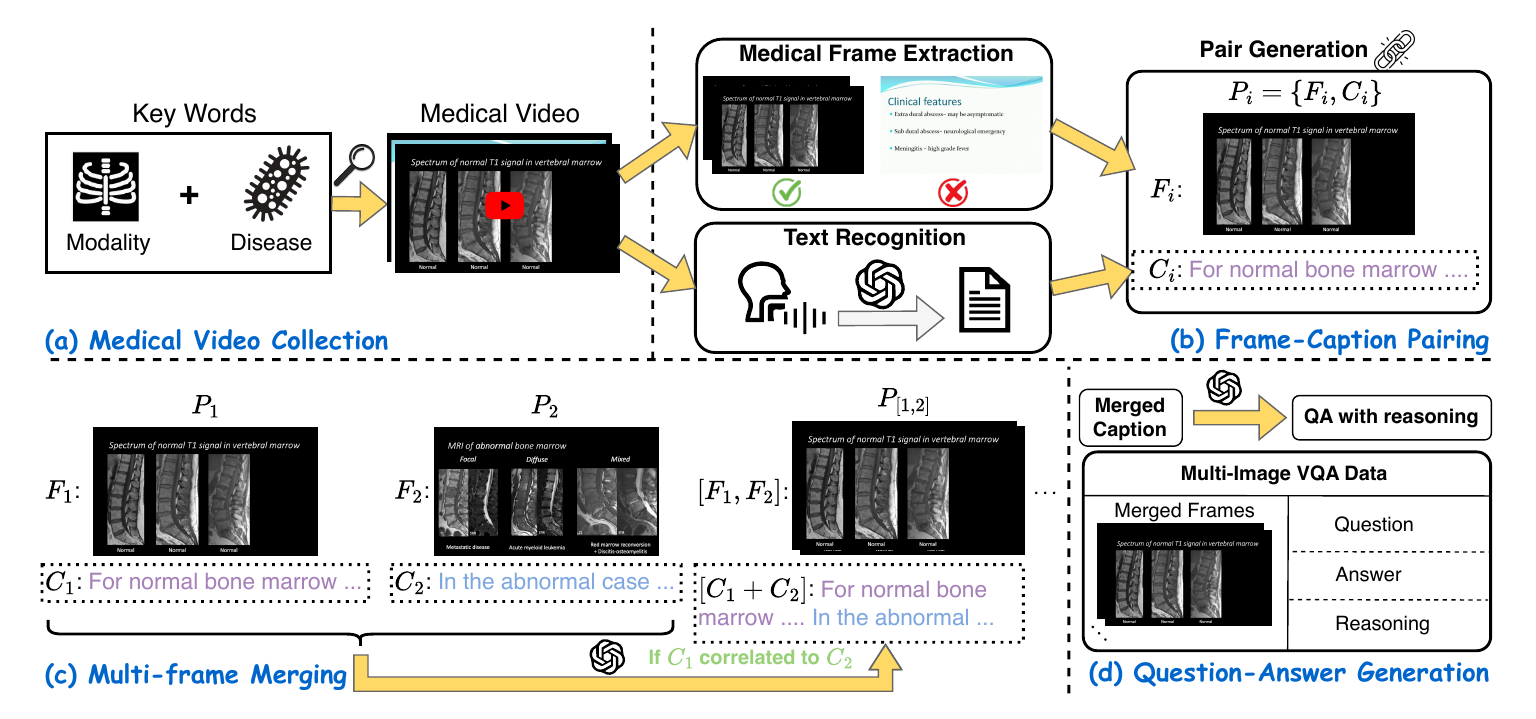}
  \vspace{-2em}
  \caption{\textbf{Data generation pipeline.}
  (a) Video Collection: Collecting 3,420 medical videos via clinical search queries (\cref{video_collection}). (b) Frame-Caption Pairing:  Extracting keyframes and aligning with transcribed captions. (\cref{pair_process}) (c) Multi-Frame Merging: Merging clinically related frame-caption pairs into multi-frame clips. (\cref{frame_merge})(d) Question-Answer Generation: Generating multi-image VQA from the multi-frame clips. (\cref{QA_generation})
  }
  \vspace{-1em}
  \label{fig:pipeline}
\end{figure*}

As the first step in building \OurBench, we assemble a large pool of clinically relevant videos from YouTube (illustrated in \cref{fig:pipeline}(a)). 
Specifically, we curate 114 carefully designed search queries, each formed by pairing a common imaging modality (\eg MRI, X-Ray, CT, and radiograph) with a frequently encountered disease or finding (\eg brain tumor, pneumonia, chest, and bone fracture). 
This combinatorial list gives broad coverage of routine diagnostic scenarios; the full set of keywords is provided in \cref{sec:keyword_list}. 
Then, for every query, we retrieve the top related results and discard clips shorter than 5 minutes or longer than 2 hours. 
The remaining corpus comprises 1,971 high-resolution, narration-rich medical videos that serve as the raw material for \OurBench.

\subsection{Frame-Caption Pairing}\label{pair_process}

\paragraph{Medical Frame Extraction.}\label{frame_extraction}
To process the raw video collected, the first task is to identify the corresponding medical frames.
Following \citet{DBLP:conf/nips/IkezogwoSGGMAKS23}, we run FFmpeg (\href{https://ffmpeg.org/}{https://ffmpeg.org/}) to extract key-frames---those delineating the scene boundaries and often indicating significant visual transitions---and record the corresponding temporal span of each segment $(f_{\text{start}},f_{\text{end}})$.
Each candidate frame is then evaluated by GPT-4o~\citep{DBLP:journals/corr/abs-2410-21276} under four criteria: (1) \textit{image quality}, evaluating the clarity and medical relevance of the frame; (2) \textit{prominence of medical content}, determining if the frame predominantly consists of medical imagery; (3) \textit{informative content}, checking if the frame is understandable and holds significant information; and (4) \textit{privacy}, ensuring the frame excludes unrelated human faces, such as those of presenters in video conferences.
Note that only frames satisfying all four requirements are retained.  More details about the frame filtering criteria can be found in~\cref{sec:prompt_details}.

This filtering step leaves us with a sequence of qualified key-frames and their temporal spans:
\begin{equation}
\begin{split}
    S_F &= [F_1,\cdots F_m], \\
    D_F &= [\left(f_{start}^1, f_{end}^1\right),\cdots \left(f_{start}^m, f_{end}^m\right)],
\end{split}
\end{equation}
where $m$ is the number of extracted medical frames. $S_F$ and $D_F$ are the sequence of frames and times.

\paragraph{Text Recognition.}
We next transcribe the audio track with Whisper~\citep{DBLP:conf/icml/RadfordKXBMS23}.
The model returns a sequence of $n$ text snippets and their time stamps:
\begin{equation}
\begin{split}
  S_T &= [T_1,\cdots T_n], \\
  D_T &= [\left(t_{start}^1, t_{end}^1\right),\cdots \left(t_{start}^n, t_{end}^n\right)].
\end{split}
\end{equation}

\paragraph{Pair Generation.}
Our third task now is to pair the medical frame with the corresponding caption.
Intuitively, each frame can be simply paired with the text snippets that emerge concurrently with it during the same time interval.
However, narration in medical videos can lag behind or precede the exact moment a frame is shown. 
To associate each frame ($F_i$) with all relevant speech, we define a symmetric margin ($\Delta$) seconds around the frame’s interval and gather every transcript whose span intersects that window $\bigl[f_{\text{start}}^{i}-\Delta,f_{\text{end}}^{i}+\Delta\bigr]$.
Then all snippets within this window range will be concatenated to form a coarse caption
$\tilde{C}_i
= \bigl[T_{j},T_{j+1},\dots,T_{k}\bigr]$.

\begin{table*}[t]
  \centering
  \small
  \begin{tabular}{lcccccccccc}
    \toprule
    \raisebox{-1ex}{\textbf{Model}} &
    \multicolumn{9}{c}{\textbf{Accuracy per System}} &
    \raisebox{-1ex}{\textbf{Avg}} \\
    \cmidrule(lr){2-10}
    & \textbf{CNS} & \textbf{RES} & \textbf{CIR} & \textbf{DIG} & \textbf{URI} & \textbf{REP} & \textbf{END} & \textbf{MSK} & \textbf{AUX} & \\
    \midrule
    \multicolumn{11}{c}{\textit{Proprietary Reasoning Models}} \\
    \midrule
    o1 & 46.91 & 48.88 & 49.49 & 47.45 & 49.03 & 42.26 & 47.68 & 51.59 & 48.75 & 47.91 \\
    o3 & 47.81 & 52.00 & 50.00 & 48.48 & 50.71 & 45.02 & 51.84 & 54.90 & 50.41 & 50.18 \\
    o4-mini & 46.03 & 49.78 & 48.74 & 48.63 & \textbf{51.85} & 43.62 & 52.44 & 53.38 & 50.82 & 49.40 \\
    Gemini-2.5-Flash & 48.82 & \textbf{58.26} & \textbf{57.21} & \textbf{50.25} & 48.61 & \textbf{55.81} & \textbf{55.38} & \textbf{60.21} & \textbf{52.85} & \textbf{54.75} \\
    Claude-3.7-Sonnet & 49.21 & 46.09 & 53.23 & \textbf{50.25} & 49.07 & 47.57 & 47.81 & 52.42 & 49.59 & 49.67 \\
    \midrule
    \multicolumn{11}{c}{\textit{Open-Source Reasoning Models}} \\
    \midrule
    QvQ-72B-Preview & 44.88 & 46.67 & 47.43 & 41.13 & 45.68 & 47.00 & 47.68  & 49.37 & 47.15 & 46.44\\
    \midrule
    \multicolumn{11}{c}{\textit{Proprietary Non-Reasoning Models}} \\
    \midrule
    GPT-4o & 48.82 & 49.13 & 37.31 & 50.00 & 43.98 & 45.88 & 46.22 & 43.60 & 44.31 & 45.67 \\
    GPT-4o-mini & 41.73 & 36.52 & 39.30 & 28.36 & 35.65 & 33.83 & 30.68 & 34.95 & 34.96 & 34.55 \\
    GPT-4-Turbo-V & 45.28 & 46.09 & 42.79 & 49.75 & 43.06 & 48.63 & 49.80 & 45.16 & 46.75 & 46.69\\
    \midrule
    \multicolumn{11}{c}{\textit{Open-Source Non-Reasoning Models}} \\
    \midrule
    Qwen2.5-VL-72B-Instruct & 43.18 & 47.39 & 42.29 & 39.80 & 39.81 & 43.41 & 43.03 & 44.00 & 40.11 & 42.65 \\
    \midrule
    \multicolumn{11}{c}{\textit{Open-Source Non-Reasoning Medical Finetuned Models}} \\
    \midrule
    MedGemma-27b-it & \textbf{49.61} & 44.20 & 48.09 & 43.45 & 41.36 & 46.58 & 50.33 & 45.62 & 39.70 & 45.47\\
    \bottomrule
  \end{tabular}
  \caption{\textbf{Accuracy of Models on \OurBench.} We report the system-wise accuracy of models on \OurBench.
  The results are averaged over all the tasks in \OurBench.
  The best results on each system and average accuracy are highlighted in bold.
  In general, all assessed models demonstrate persistently low accuracy, with system-wise performance of substantial variability in task difficulty.
  }\label{tab:results}
  \vspace{-1.5em}
\end{table*}

Then we leverage \texttt{GPT-4o} to enhance the quality of $\tilde{C}_i$. 
Specifically, \texttt{GPT-4o} is instructed to 
(i) remove statements unrelated to the displayed frame and
(ii) refine the description to ensure the correct usage of clinical terminology.
Formally,
\begin{equation}
    C_i = \texttt{GPT-4o}\left(\tilde{C}_i, F_i \mid I_{rephrase}\right),
\end{equation}
where $C_i$ denotes the refined caption, and $I_{rephrase}$ is the prompt (see \cref{sec:prompt_details} for more details).
The final frame–caption pair is $P_i=\{F_i,C_i\}$, and the sequence of frame-caption pairs of the entire video is $S_P = [P_1,\cdots,P_n]$.

\subsection{Multi-Frame Merging}\label{frame_merge}
The paired frames described above usually belong to longer narrative units within educational presentations---for example, a radiologist may spend several consecutive slides discussing the same lesion during a structured teaching session. 
To capture such continuity, we merge adjacent frame-caption pairs into multi-frame ''clips'' whenever their captions describe the same clinical concept within the educational context.
The paired caption of each frame already provides a description of its visual content; hence, we rely entirely on the textual correlation between the captions to determine if there is a connection between two frames.
Specifically, as illustrated in~\cref{fig:pipeline}(c), for every consecutive pair
$P_i=\{F_i,,C_i\}$ and $P_{i+1}=\{F_{i+1},,C_{i+1}\}$,
we ask \texttt{GPT-4o} (prompt in \cref{sec:prompt_details}) whether these two captions are correlated. 
If yes, we then combine these two pairs:
$P_{[i,i+1]} = \bigl\{[F_i,F_i+1], [C_i \oplus C_{i+1}]\bigr\}$,
where $\oplus$ represents the text concatenation.
We then compare the merged caption $[C_i \oplus C_{i+1}]$ with the next caption $C_{i+2}$; if the relation persists, we append $P_{i+2}$ to the group.  This sliding process continues until (i) the next caption is judged unrelated or (ii) the group reaches a maximum of five frames, the limit we adopt in this work.

Applying the above procedure to all videos yields 7,998 multi-frame clips, each containing 2–5 medically coherent frame-caption pairs.  
These clips constitute the basic building blocks for the subsequent VQA-item generation stage.

\subsection{Question Answering Generation}\label{QA_generation}
As shown in~\cref{fig:pipeline}(d), for each merged group $P_{[i,i+1\cdots]} = \{[F_i,F_{i+1},\cdots],[C_i\oplus C_{i+1},\cdots]\}$, we instruct \texttt{GPT-4o} to generate challenging multiple-choice questions. Formally,
\begin{equation}
    Q,A,R = \texttt{GPT-4o}\left([C_i\oplus C_{i+1}\cdots]\mid I_{gen}\right),
\end{equation}
where $Q,A,R$ are the generated question, the correct answer, and the reasoning, respectively. 
$I_{gen}$ is the generation prompt, enforcing four requirements: (1) \textit{Information Grounding}: all questions must rely solely on visual evidence explicitly described in the educational video captions; (2) \textit{Educational Clinical Reasoning}: each question should probe skills demonstrated in medical education contexts such as anatomical localization and differential diagnosis within structured presentations; (3) \textit{Contextual Interaction}: the wording must reference the images in order (\eg, ``in the first image ..., whereas in the third image ...'') and require synthesizing information across the educational sequence; (4) \textit{Distraction Options}: every item includes plausible but incorrect answer choices that differ from the ground truth in clinical  details within the educational context.
The complete $I_{gen}$ is provided in \cref{sec:prompt_details}.
Lastly, each clip is packaged as $\{Q,A,R,[F_i,F_{i+1}\cdots]\}$, forming a single entry.

\begin{table*}[t]
  \centering
  \footnotesize
  \setlength{\tabcolsep}{4pt}
  \begin{tabular}{lcccccccccc}
      \toprule
      \multirow{2}{*}{\textbf{Model}} &
      \multicolumn{5}{c}{\textbf{Accuracy (\%) by Frame Count}} &
      \multicolumn{5}{c}{\textbf{Accuracy (\%) by Modality}} \\
      \cmidrule(lr){2-6} \cmidrule(lr){7-11}
      & 2 & 3 & 4 & 5 & \textit{SD} & CT & MRI & Ultrasound & X-ray & Other \\
      \midrule
      o1 & 48.16 & 45.64 & 51.43 & 48.15 & 2.37 & 48.98 & 45.40 & 49.05 & 49.16 & 51.64 \\
      o3 & 50.00 & 47.46 & 53.60 & 51.38 & 2.57 & 50.09 & 48.57 & 51.45 & 53.06 & 52.38 \\
      o4-mini & 50.21 & 46.23 & 50.00 & 50.37 & 1.99 & 48.08 & 48.85 & 52.34 & 50.33 & 53.49 \\
      Gemini-2.5-Flash & \textbf{53.54} & \textbf{55.48} & \textbf{55.47} & \textbf{55.76} & 1.02 & \textbf{54.57} & \textbf{53.60} & \textbf{57.36} & \textbf{58.14} & 49.24 \\
      QvQ-72B-Preview & 48.00 & 46.73 & 42.32 & 45.23 & 2.12 & 45.18 & 47.62 & 48.32 & 44.08 & 47.98 \\
      
      GPT-4-Turbo-V & 47.47 & 45.51 & 46.88 & 46.34 & 0.83 & 46.83 & 43.48 & 50.65 & 49.17 & 51.52 \\
      GPT-4o & 47.30 & 45.18 & 40.23 & 45.35 & 3.01 & 45.52 & 43.27 & 48.58 & 47.51 & 51.52 \\
      GPT-4o-mini & 35.16 & 36.21 & 32.42 & 33.09 & 1.77 & 35.26 & 34.31 & 34.88 & 34.55 & 29.55 \\
      Claude-3.7-Sonnet & 49.41 & 48.01 & 51.56 & 50.68 & 1.55 & 50.75 & 49.11 & 49.10 & 49.83 & 46.21 \\
      Qwen2.5-VL-72B-Instruct & 42.72 & 41.14 & 42.71 & 43.66 & 0.90 & 40.95 & 43.52 & 42.64 & 45.07 & 44.70 \\
      MedGemma-27b-it & 43.73 & 44.80 & 46.88 & 48.08 & 1.70 & 47.64 & 43.03 & 44.10 & 43.19 & \textbf{54.08} \\
      \bottomrule
  \end{tabular}
  \caption{\textbf{Accuracy (\%) of Models by Frame Count and Modality on \OurBench.}
  We report the accuracy of models on questions in \OurBench grouped by frame count with standard deviation (\textit{SD}) and by modality.
  We empirically observe that accuracy fluctuates with increasing frame count and varies significantly across common imaging modalities.}
  \label{tab:results_per_frame}
\end{table*}

\subsection{Data Filtering}
\label{data_filtering}
\textbf{Difficulty Filtering.} 
To ensure difficulty, we filtered using GPT-4-Turbo-V~\citep{GPT4V}, \texttt{o1}~\citep{DBLP:journals/corr/abs-2412-16720}, and \texttt{GPT-4o}~\citep{DBLP:journals/corr/abs-2410-21276}. We discarded items where all models reached a consensus (either \emph{all} correct or \emph{all} identically incorrect) to eliminate trivial or potentially erroneous keys.
This step trims the pool from 4,457 to 3,654 items.

\textbf{Human Evaluation.} We manually filtered the dataset to ensure quality and privacy, excluding 803 entries containing blurred frames, recognizable faces, or insignificant medical content. This yielded final 2,851 high-quality entries.
\section{Experiments}
\label{experiments}
\definecolor{darkcolor}{HTML}{99CCFF}
\definecolor{shallowcolor}{HTML}{CCE5FF}

\begin{figure*}[t]
  \centering
  \includegraphics[width=\linewidth]{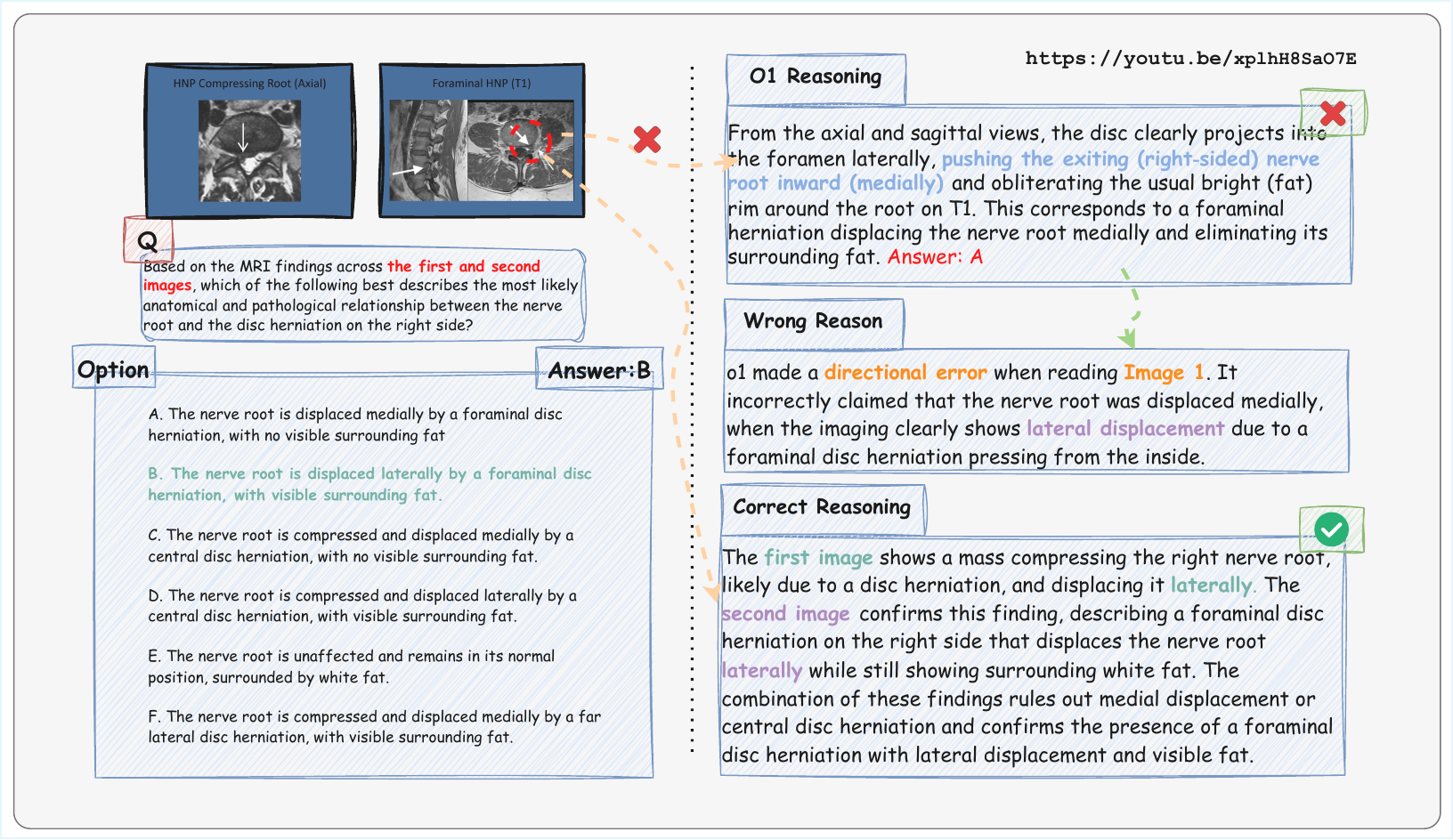}
  \caption{\textbf{Failure case study of \texttt{o1} on \OurBench.} A single image mistake can cause significant errors. Here, \texttt{o1} made a directional error when interpreting the first frame, which propagated through its reasoning process and ultimately led to an incorrect answer.}
  \vspace{-1em}
  \label{fig:failure_case_study}
\end{figure*}

\subsection{Data Statistics}\label{sec:data_dist}
In this section we summarize the data distribution of \OurBench.
Starting from the 3,420 instructional videos collected in \cref{video_collection}, we extract 111,942 key-frames and retain 9,237 high-quality, medically relevant frames.
These frames are used to construct 2,851 multi-image, closed-ended, single-choice VQA pairs, which span 9 human body systems and 43 organs, featuring 114 unique keyword combinations following \citet{herring2019learning}. 
Each generated VQA pair consists of 2–5 frames, accompanied by a challenging question that requires integrating information across all provided frames to answer correctly.
The composition of body systems, organs and modalities in \OurBench is provided in \cref{appendix:Data Distribution} and shown in \cref{fig:data_distribution} (a) (b) (c) respectively.

A defining feature of \OurBench is that every question is tethered to multiple images, deliberately pushing models to reason across frames—a core requirement in real-world diagnosis. 
Concretely, among the 2,851 VQA items,
1,186 pairs contain 2 frames, 602 pairs contain 3 frames, 256 pairs contain 4 frames, and 807 pairs contain 5 frames.
We also present the distribution of frames per question in \cref{fig:data_distribution}(e).

\subsection{Models}
\label{evaluation_models}
We evaluate both proprietary and open-source MLLMs on \OurBench, encompassing reasoning and non-reasoning models, with a particular focus on recent advancements in medical reasoning. For evaluation, we use the prompt template as in MMMU-pro\citep{DBLP:journals/corr/abs-2409-02813} (see~\cref{sec:prompt_details}).

\textbf{Reasoning Models:} We evaluate both proprietary reasoning model and open-source reasoning model, including 4 proprietary models \texttt{o4-mini} \citep{openai2025o3o4}, \texttt{o3} \citep{openai2025o3o4}, \texttt{o1} \citep{DBLP:journals/corr/abs-2412-16720}, \texttt{Claude-3.7-Sonnet} \citep{anthropic2025} and \texttt{Gemini-2.5-Flash} \citep{gemini2025} and one open-source \texttt{QvQ-72B-Preview} \citep{team2024qvq}.

\textbf{Non-Reasoning Models:} We evaluate \OurBench on non-reasoning models. including 3 proprietary models, \texttt{GPT-4o} \citep{DBLP:journals/corr/abs-2410-21276}, \texttt{GPT-4o-mini} \citep{DBLP:journals/corr/abs-2410-21276} and \texttt{GPT-4-Turbo-V} \citep{GPT4V} and 2 open-source models \texttt{Qwen2.5-VL-72B-Instruct} \citep{DBLP:journals/corr/abs-2502-13923} and the medical fine-tuned model \texttt{MedGemma-27b-it} \citep{DBLP:journals/corr/abs-2507-05201} to evaluate domain-specific adaptations.

\subsection{Main Results}
\paragraph{Advanced MLLMs struggle to holistically understanding multi-images.}
Table \ref{tab:results} presents the evaluation of 11 advanced MLLMs on \OurBench. In general, all assessed models demonstrate persistently low accuracy, with the peak accuracy remaining below 55.00\%.

The proprietary model, \texttt{GPT-4o}, reaches an average accuracy of 45.67\%, significantly lower in comparison to its performance on the single medical VQA benchmark (69.91\% on VQA-RAD~\citep{lau2018dataset} as reported by~\citet{yan2024worse}).
The leading open-source model, \texttt{Qwen2.5-VL-72B-Instruct}, achieves merely 42.65 ± 0.34\% (SE) accuracy.
To confirm this stemmed from reasoning deficits rather than inadequate medical knowledge, we evaluated \texttt{MedGemma-27b-it}, which similarly yielded poor results with 45.47 ± 0.59\% (SE) accuracy.

\paragraph{Reasoning enhances multi-image understanding.}
As shown in \cref{tab:results}, we find that reasoning MLLMs consistently outperform non-reasoning ones. 
\texttt{Gemini-2.5-Flash} attains the highest accuracy among all models, notably outperforming the top non-reasoning model \texttt{GPT-4o} by 9.08\% (54.75\% \textit{vs} 45.67\%).
Among the open-source models, \texttt{QvQ-72B-Preview} achieves an accuracy of 46.44\% ± 0.66\% (SE), showcasing a 3.79\% enhancement compared to its non-reasoning counterpart, \texttt{Qwen2.5-VL-72B-Instruct}.
This indicates that reasoning is particularly beneficial in clinical scenarios, which frequently involve interpreting multiple images.


\begin{figure*}[t]
  \centering
  \includegraphics[width=\linewidth]{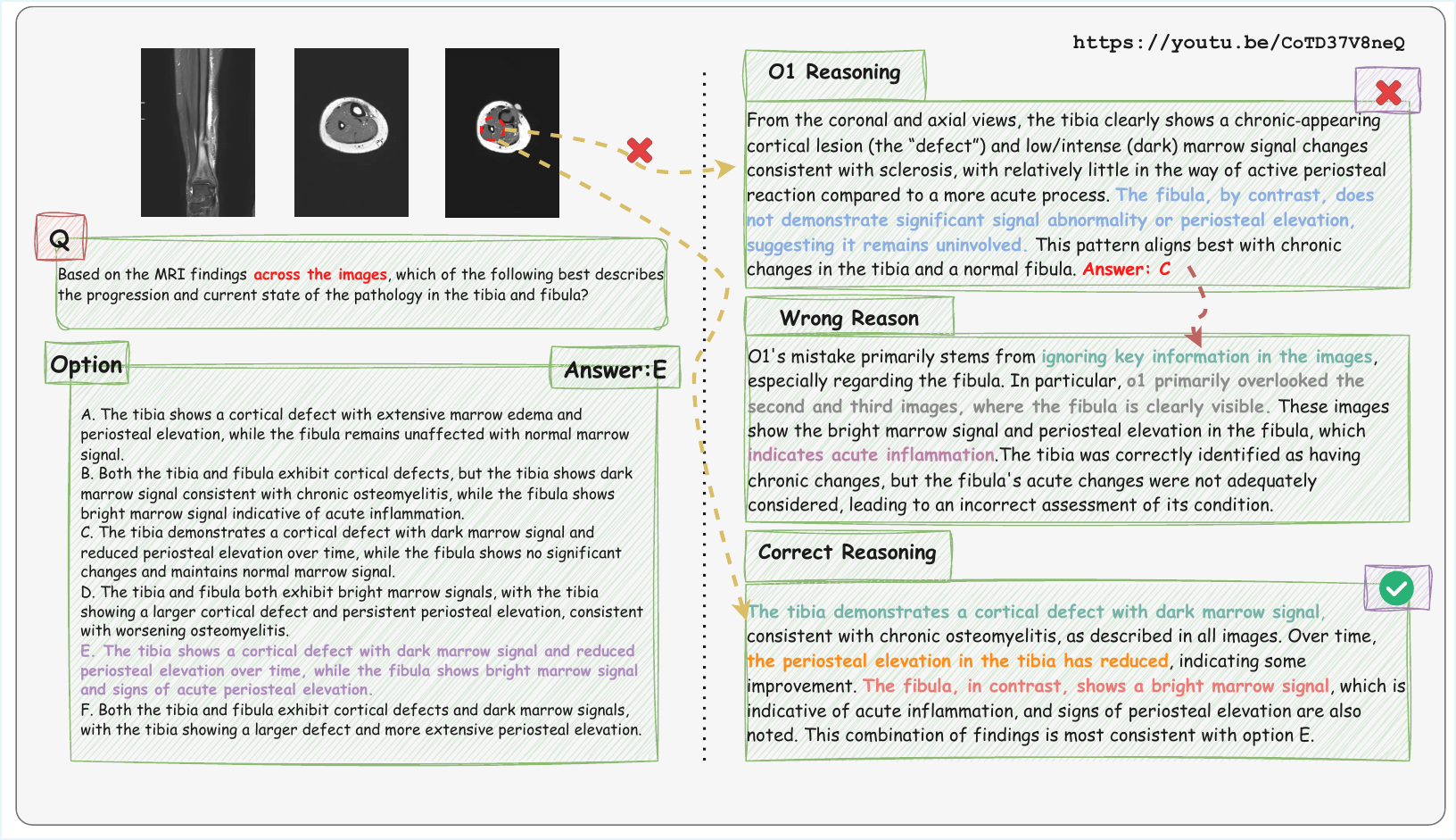}
  \vspace{-1em}
  \caption{\textbf{Failure case study of \texttt{o1} on \OurBench.} Negligence of important information across multiple frames. In this case, \texttt{o1} overlooked critical features in the second and third frames, which ultimately led to the selection of an incorrect answer.}
  \label{fig:failure_case_study_2}
  \vspace{-1em}
\end{figure*}

\paragraph{Overlooking or misinterpreting hinders reasoning across image sequence.} 
Despite the relatively enhanced performance of reasoning models, their performance is still limited.
Our investigation reveals this primarily arises from neglecting or misinterpreting the intermediary images during continuous reasoning over an image sequence.
Here, we present a case study highlighting instances where \texttt{o1} fails to provide correct reasoning steps for questions in \OurBench:

\emph{Case 1: Neglegence of important information within multiple frames.}
In \cref{fig:failure_case_study_2}, \texttt{o1} fails to integrate important information across multiple frames, leading to a flawed overall reasoning.
While \texttt{o1} correctly identifies the “polar vessel sign” in the Doppler frame as suggestive of a parathyroid adenoma, 
it ignores distinct transverse and sagittal localization cues (posterior–inferior to the thyroid with specific cranial–caudal orientation), leading to an incorrect conclusion.

\emph{Case 2: Mistake drawn from single image resulting in significant errors in subsequent reasoning.}
In \cref{fig:failure_case_study}, \texttt{o1} misreads a critical axial frame, incorrectly identifying medial rather than lateral nerve root displacement caused by a foraminal disc herniation. This error propagates through the reasoning process, yielding an anatomically incorrect answer that conflicts with evidence from both frames.

\subsection{Evaluation across anatomical structures or frame numbers}

\textbf{Comparisons between anatomical structures and modalities.} 
The system-wise performance we report in \cref{tab:results} reveals substantial variability in task difficulty.
For instance, \texttt{Gemini-2.5-Flash} achieves an accuracy of 60.21\% on questions related to the musculoskeletal system, but only 48.61\% on the urinary system, resulting in 11.60\% gap.
In \cref{appendix:comparison_of_organs}, we present a detailed analysis of performance variation across four representative organs in \OurBench.
We also report the performance of MLLMs across different imaging modalities in \cref{tab:results_per_frame}.
Notably, the accuracy varies significantly across common modalities such as CT, MRI, Ultrasound, and X-ray.
\texttt{QvQ-72B-Preview} exhibits a 4.24\% performance gap between Ultrasound and X-ray, whereas \texttt{Gemini-2.5-Flash} shows a 4.54\% gap between MRI and X-ray, underscoring the strong modality sensitivity of current MLLMs and indicating the need for more diverse and balanced modality–organ combinations during training to improve generalization.

\textbf{Comparisons betweem VQAs with different numbers of frames.}
In \cref{tab:results_per_frame}, we report the accuracy of models on questions in \OurBench, grouped by the number of frames each question contains.
Empirically, we observe that accuracy fluctuates as the number of images per question increases, with performance improving at certain frame counts and declining at others. 
Among the MLLMs, \texttt{GPT-4o} exhibits substantial fluctuation, with a standard deviation of 3.01, whereas \texttt{GPT-4-Turbo-V} shows minimal variation, with a standard deviation of just 0.83. 
These fluctuations suggest that performance depends less on the number of frames than on the complexity or redundancy of visual information across them.
\section{Conclusion}
We introduces \OurBench, a multi-image medical visual question answering benchmark, comprising 2851 multi-image questions, sourced from 3420 medical videos of 114 keywords and covering over 43 organs. 
We propose an automated pipeline to generate high-quality multi-image VQA data ensuring semantic progression and contextual consistency across frames. 
Unlike existing single-image datasets, \OurBench has both multi-image question answering pairs and a detailed reasoning process, containing 2-5 images input and 3.24 images input per question. 
We comprehensively benchmark ten state-of-the-art models, presenting accuracies predominantly below 50\%.
We hope \OurBench paves the way for future multi-modal medical reasoning research.


\section*{Limitations} \label{Limitation}
A key limitation is that \OurBench has not received full-scale physician evaluation. We only obtained clinician review for a subset of questions, and the feedback on this subset indicated the questions are overall quite good. Nevertheless, broader expert assessment is needed to further verify clinical correctness and coverage.
While \OurBench reveals clear evidence of current MLLMs’ inability in handling multi-image questions of clinical reasoning, effective strategies to enhance their multi-image reasoning capabilities remain underexplored. Future work will focus on developing and evaluating methods to improve such capabilities. We believe \OurBench will serve as a valuable resource for advancing research in multimodal medical AI and fostering the development of more capable and robust diagnostic reasoning systems.

\section*{Ethical Considerations}
The MedFrameQA benchmark was constructed exclusively using publicly available medical education videos hosted on YouTube. This study did not involve the recruitment of human subjects, direct interaction with patients, or the collection of private clinical data. To strictly uphold data privacy and ethical standards, a comprehensive multi-stage filtering protocol was implemented. All extracted video frames were subjected to both automated screening via GPT-4o and rigorous manual review to ensure the complete removal of personally identifiable information. Specifically, any frames containing recognizable human faces, including those of patients or video presenters, were excluded from the dataset. The final dataset consists solely of de-identified medical imagery derived from open-access educational content, ensuring compliance with privacy norms while fostering the development of robust clinical AI systems.

\section*{Acknowledgments}
We thank the Microsoft Accelerate Foundation Models Research Program for supporting our computing needs.


\bibliography{references}

\appendix
\appendix
\newpage


\section{Use of LLMs}
We employed large language models (LLMs) in the dataset construction pipeline to refine and filter captions, identify and merge semantically related captions, and generate multi-image VQA items. 
We further benchmarked state-of-the-art MLLMs on MedFrameQA.

During the preparation of this manuscript, we used OpenAI’s GPT-4.1 model for minor language refinement and smoothing of the writing. 
The AI tool was not used for generating original content, conducting data analysis, or formulating core scientific ideas. 
All conceptual development, experimentation, and interpretation were conducted independently without reliance on AI tools.

\section{Data Distribution}\label{appendix:Data Distribution}
\begin{figure*}[p]
  \centering
  \vspace*{\fill}
  \includegraphics[width=\textwidth,height=0.85\textheight,keepaspectratio]{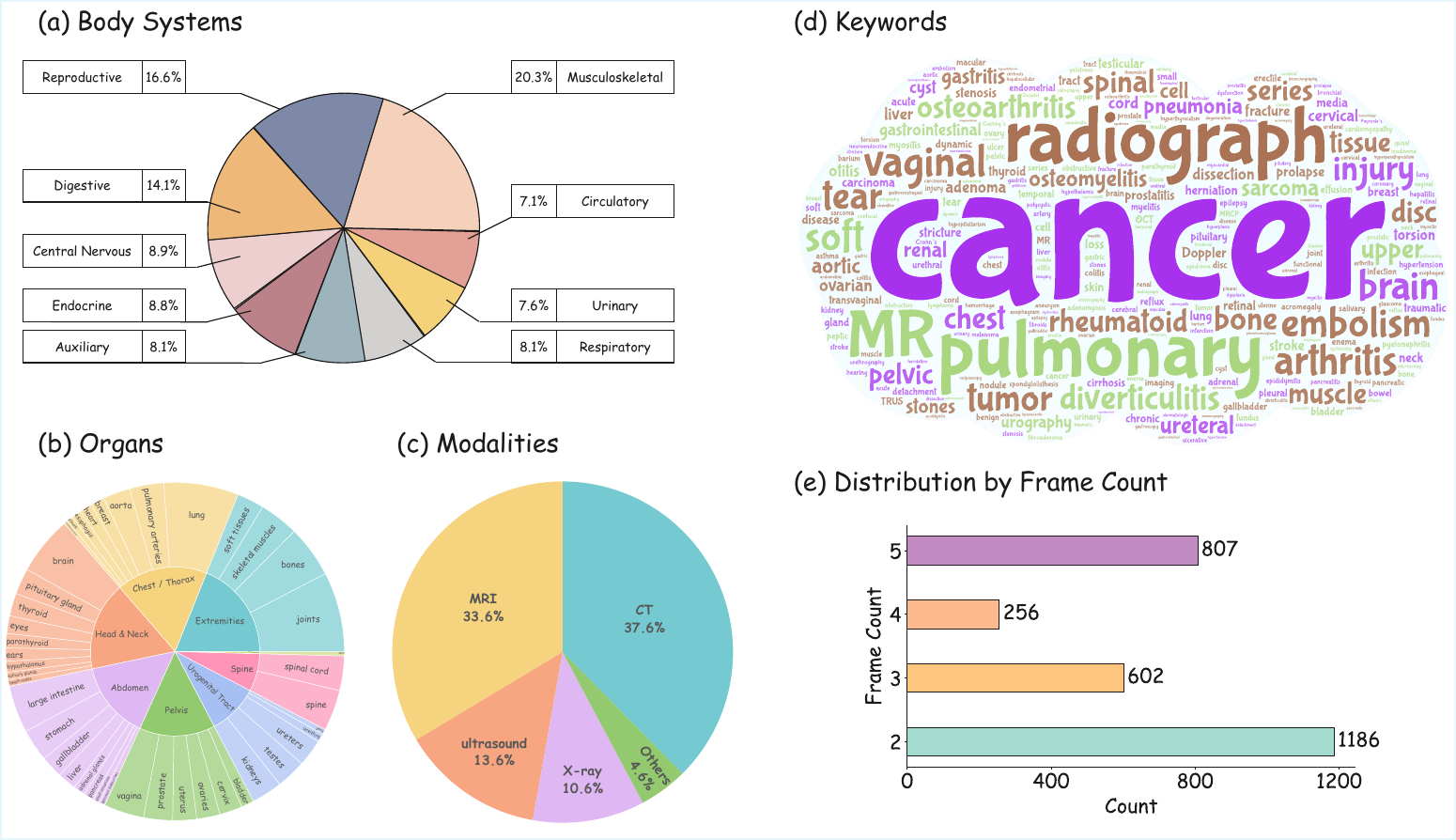}
  \vspace*{\fill}
  \caption{\textbf{Data distribution of \OurBench.}
  In \cref{fig:data_distribution}(a), we show the distribution across body systems;
    (b) presents the distribution across organs;
    (c) shows the distribution across imaging modalities;
    (d) provides a word cloud of keywords in \OurBench;
    and (e) reports the distribution of frame counts per question.
  }
  \label{fig:data_distribution}
\end{figure*}

We present detailed data distributions across body systems, organs, and imaging modalities in \cref{fig:data_distribution}(a), (b), and (c), respectively.
A word cloud of keywords in \OurBench is shown in \cref{fig:data_distribution}(d), and the distribution of frame counts per question is provided in \cref{fig:data_distribution}(e).

\section{API Cost}\label{api cost}
Generation of each data entry costs 5 times calling of \texttt{GPT-4o} API on average, depending on the number of frames involved in the data entry.
Construction of 2,851 data entries costs 14,255 API calls in total.

For proprietary models (e.g., \texttt{GPT-4o}, \texttt{Gemini-2.5-Flash}, \texttt{Claude-3.7-Sonnet}), we use their official APIs and perform 2,851 requests per model, corresponding to the number of examples in \OurBench.

For open-source models (e.g., \texttt{QvQ-72B-Preview}, \texttt{Qwen2.5-VL-72B-Instruct}, \texttt{MedGemma-27b-it}), we conducted three independent runs on 4×A100 GPUs and calculated error bars. Due to API quota constraints, proprietary models were evaluated only once.




\newpage
\section{Keyword List} \label{sec:keyword_list}
\begin{figure*}[p]
  \centering
  \vspace*{\fill}
  \includegraphics[width=\textwidth,height=0.95\textheight,keepaspectratio]{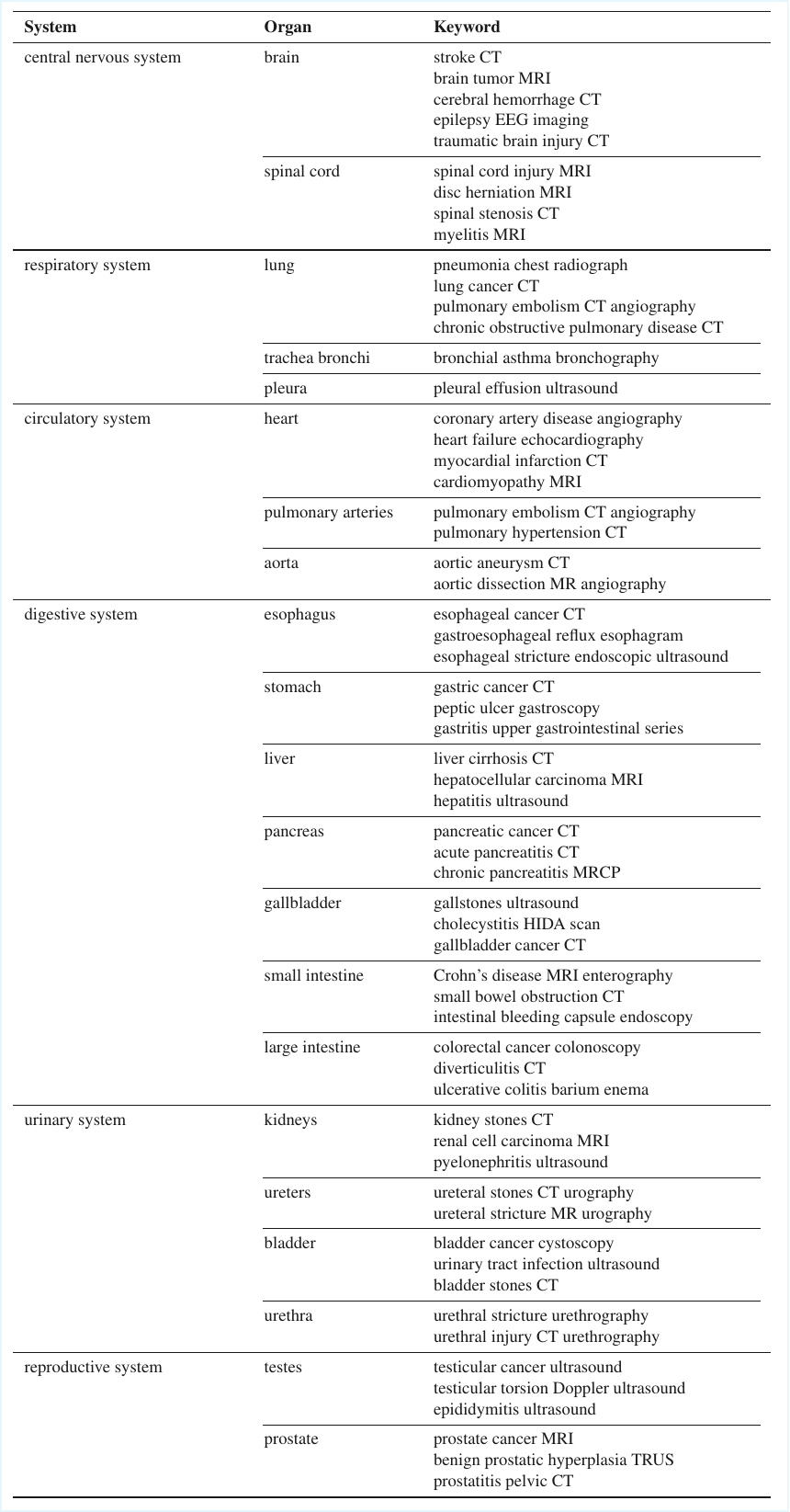}
  \caption{Keyword List (Part 1)}
  \label{fig:keyword_list_1}
  \vspace*{\fill}
\end{figure*}

\begin{figure*}[p]
  \centering
  \vspace*{\fill}
  \includegraphics[width=\textwidth,height=0.95\textheight,keepaspectratio]{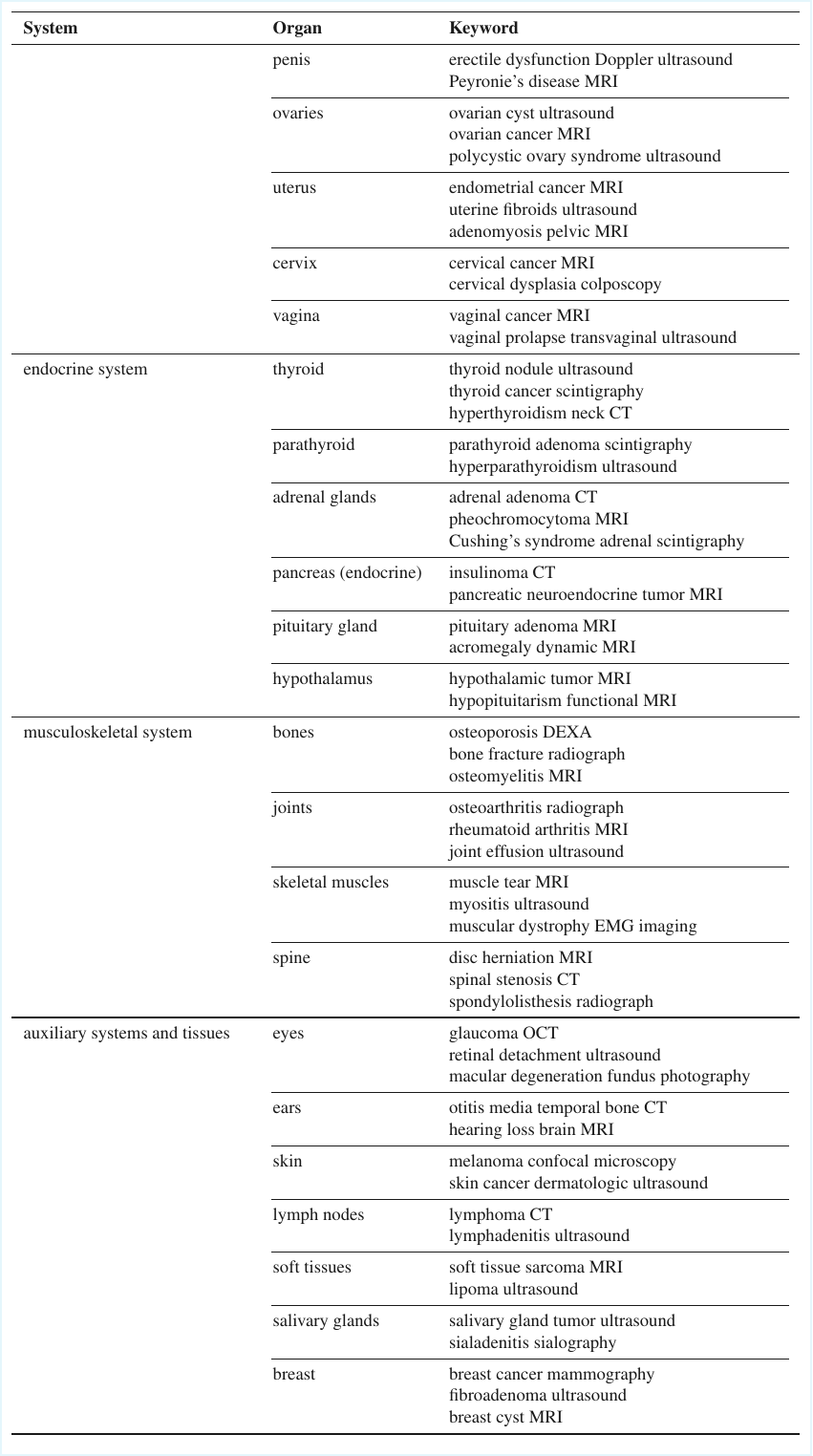}
  \par\medskip
  \caption{Keyword List (Part 2)}
  \label{fig:keyword_list_2}
  \vspace*{\fill}
\end{figure*}

We present comprehensive list of search queries utilized for video collection.
As detailed in \cref{video_collection}, we curated a total of 114 combinatorial search queries to ensure broad coverage of routine diagnostic scenarios.
Each query is formed by pairing a specific imaging modality (e.g., MRI, X-Ray, CT) with a frequently encountered clinical disease.
These keywords, listed in Figure \ref{fig:keyword_list_1} and Figure \ref{fig:keyword_list_2}, span 9 human body systems and 43 organs, serving as the foundational criteria for retrieving high-quality medical educational videos from YouTube.
\clearpage
\newpage
\section{Comparison of Organs} \label{appendix:comparison_of_organs}
We present a detailed organ-wise accuracy comparison of 11 state-of-the-art MLLMs on \OurBench in \cref{tab:comparison_of_organs}.
Our results reveal substantial performance variation across different organs.
While \texttt{Gemini-2.5-Flash} outperforms other models on average in \cref{tab:results}, open-source models like \texttt{QvQ-72B-Preview} demonstrate competitive performance on specific organs, such as the ureters and pulmonary arteries.
This variability highlights the sensitivity of MLLM performance to the anatomical structures involved, underscoring the need to develop models that are more robust to anatomical diversity.
This variability underscores the sensitivity of MLLM performance to organ-specific features and highlights the need for future research focused on improving anatomical generalization across a wide range of clinical scenarios.

\begin{table*}[p]
    \centering
    \vspace*{\fill}
    \scriptsize
    
    \setlength{\tabcolsep}{1pt}
    \begin{tabular}{l*{11}{>{\centering\arraybackslash}p{1.2cm}}}
      \toprule
      \multicolumn{1}{c}{\raisebox{-1.5ex}{\textbf{Organs}}} &
      \multicolumn{11}{c}{\textbf{Model Accuracy}} \\
      \cmidrule(lr){2-12}
      & Gemini-2.5-Flash & Claude-3.7-Sonnet & o4-mini & o3 & o1 & GPT-4o & GPT-4o-mini & GPT-4-Turbo-V & QvQ-72B & Qwen2.5-VL-72B-Instruct & MedGemma-27b-it \\
      \midrule
      \multicolumn{11}{l}{\textbf{auxiliary systems and tissues}} \\
      \midrule
      \rowcolor{gray!15}
      soft tissues          & \textbf{48.65}	& 37.84	& 45.95	& 39.19	& 35.14	& 36.49	& 32.43	& 35.14	& 40.54	& 30.63 & 35.68 \\
      salivary glands       & 55.00	& 50.00	& 45.00	& 52.63	& 47.37	& 40.00	& 40.00	& 45.00	& \textbf{66.67}	& 48.33 & 43.33\\
      \rowcolor{gray!15}
      skin                  & 33.33	& 66.67	& 50.00	& 70.00	& 54.55	& \textbf{75.00}	& 41.67	& \textbf{75.00}	& 36.11	& 63.89 & 50.00\\
      breast                & 52.63	& 55.26	& 55.26	& 57.89	& \textbf{58.33}	& 42.11	& 39.47	& 39.47	& 50.88	& 35.09 & 41.23\\
      \rowcolor{gray!15}
      lymph nodes           & 61.11	& \textbf{77.78}	& 72.22	& 72.22	& 61.11	& 55.56	& 27.78	& 61.11	& 53.70	& 55.56 & 53.70\\
      ears                  & \textbf{58.33}	& 47.22	& 44.44	& 52.78	& 57.14	& 50.00	& 30.56	& 55.56	& 46.30	& 37.04 & 40.74\\
      \rowcolor{gray!15}
      eyes                  & \textbf{56.25}	& 50.00	& 54.17	& 46.81	& 51.06	& 43.75	& 37.50	& 52.08	& 47.22	& 45.83 & 36.11\\
      \midrule
      \multicolumn{11}{l}{\textbf{central nervous system}} \\
      \midrule
      \rowcolor{gray!15}
      brain                 & 50.00	& 49.38	& 42.41	& 45.86	& 46.05	& 51.25	& 44.38	& 46.88	& 42.92	& 42.50 & \textbf{51.87}\\
      spinal cord           & 46.81	& 48.94	& \textbf{52.13}	& 51.06	& 48.35	& 44.68	& 37.23	& 42.55	& 48.23	& 44.33 & 45.74\\
      \midrule
      \multicolumn{11}{l}{\textbf{circulatory system}} \\
      \midrule
      \rowcolor{gray!15}
      pulmonary arteries    & 54.84	& \textbf{56.99}	& 50.54	& 49.46	& 51.09	& 43.01	& 44.09	& 47.31	& 51.97	& 44.09 & 49.82\\
      aorta                 & \textbf{60.81}	& 48.65	& 45.21	& 50.00	& 45.83	& 35.14	& 35.14	& 41.89	& 43.69	& 40.09 & 52.70\\
      \rowcolor{gray!15}
      heart                 & \textbf{55.88}	& 52.94	& 51.52	& 51.52	& 53.12	& 26.47	& 35.29	& 32.35	& 43.14	& 42.16 & 37.25\\
      \midrule
      \multicolumn{11}{l}{\textbf{digestive system}} \\
      \midrule
      \rowcolor{gray!15}
      large intestine       & 47.29	& 47.29	& 42.64	& 38.28	& 41.73	& \textbf{48.06}	& 23.26	& 46.51	& 35.14	& 31.52 & 37.98\\
      esophagus             & 59.26	& 51.85	& \textbf{70.37}	& 62.96	& 59.26	& 62.96	& 22.22	& 62.96	& 61.73	& 38.27 & 60.49\\
      \rowcolor{gray!15}
      small intestine       & 61.11	& 55.56	& \textbf{72.22}	& 58.82	& 62.50	& 44.44	& 16.67	& 55.56	& 46.30	& 50.00 & 55.56\\
      gallbladder           & 37.70	& 44.26	& 34.43	& 38.33 & 41.38	& 40.98	& 39.34	& \textbf{47.54}	& 40.98	& 36.61 & 39.34\\
      \rowcolor{gray!15}
      stomach               & 59.09	& 59.09	& 55.17	& \textbf{60.00} & 54.12	& 57.95	& 32.95	& 56.82	& 37.88	& 51.14 & 46.59\\
      liver                 & 54.90	& 54.90	& 52.94	& \textbf{60.78} & 52.94	& 50.98	& 29.41	& 43.14	& 54.25	& 46.41 & 43.14\\
      \rowcolor{gray!15}
      pancreas              & 39.29	& 35.71	& 42.86	& 39.29 & 35.71	& 42.86	& 25.00	& 42.86	& 32.14	& 32.14 & \textbf{44.05}\\
      \midrule
      \multicolumn{11}{l}{\textbf{endocrine system}} \\
      \midrule
      \rowcolor{gray!15}
      pancreas (endocrine)  & 41.18	& 35.29	& \textbf{52.94}	& 35.29 & 35.29	& 41.18	& 17.65	& 41.18	& 35.29	& 25.49 & 29.41\\
      hypothalamus          & \textbf{56.67}	& 43.33	& 53.85	& 50.00 & 42.31	& 46.67	& 43.33	& 46.67	& 45.56	& 45.56 & 52.22\\
      \rowcolor{gray!15}
      parathyroid           & 56.41	& 38.46	& 47.37	& 50.00 & 57.14	& 41.03	& 35.90	& 46.15	& 49.57	& 47.86 & \textbf{60.68}\\
      pituitary gland       & 56.34	& 56.34	& \textbf{59.15}	& 57.75 & 56.52	& 45.07	& 21.13	& 47.89	& 57.28	& 52.11 & 54.93\\
      \rowcolor{gray!15}
      adrenal glands        & \textbf{53.12}	& 43.75	& \textbf{53.12}	& 43.75 & 25.00	& \textbf{53.12}	& 40.62	& 43.75	& 41.67	& 27.08 & 45.83\\
      thyroid               & 58.06	& 51.61	& 46.77	& 55.74 & 50.00	& 48.39	& 30.65	& \textbf{61.29}	& 43.01	& 41.40 & 45.70\\
      \midrule
      \multicolumn{11}{l}{\textbf{musculoskeletal system}} \\
      \midrule
      \rowcolor{gray!15}
      spine                 & 57.14	& 49.11	& 48.21 & \textbf{58.04} & 48.65	& 47.32	& 35.71	& 50.00	& 48.81	& 46.43 & 48.51\\
      bones                 & \textbf{62.68}	& 50.70	& 51.77	& 56.83 & 54.07	& 43.66	& 37.32	& 38.03	& 55.16	& 40.38 & 41.31\\
      \rowcolor{gray!15}
      skeletal muscles      & \textbf{63.55}	& 61.68	& 62.62	& 54.29 & 50.94	& 45.79	& 38.32	& 51.40	& 50.78	& 56.39 & 57.63\\
      joints                & \textbf{58.53}	& 50.69	& 52.53	& 52.31 & 51.87	& 40.55	& 31.34	& 44.24	& 45.16	& 39.02 & 41.01\\
      \midrule
      \multicolumn{11}{l}{\textbf{reproductive system}} \\
      \midrule
      \rowcolor{gray!15}
      vagina                & \textbf{56.88}	& 50.46	& 44.44	& 47.17 & 38.24	& 49.54	& 35.78	& 54.13	& 48.01	& 43.12 & 52.60\\
      penis                 & 42.86	& 28.57	& 28.57	& 14.29 & 14.29	& 42.86	& 28.57	& 50.00	& 38.10	& \textbf{52.38} & 45.24\\
      \rowcolor{gray!15}
      ovaries               & 50.79	& 47.62	& 44.44	& 46.77 & 52.54	& 42.86	& 22.22	& 38.10	& 49.74	& \textbf{55.03} & 47.62\\
      prostate              & \textbf{50.63}	& 49.37	& 40.51	& 42.86 & 30.26	& 46.84	& 43.04	& 48.10	& 40.93	& 39.66 & 45.57\\
      \rowcolor{gray!15}
      cervix                & \textbf{61.29}	& 53.23	& 41.67	& 38.98 & 47.37	& 48.39	& 32.26	& 48.39	& 44.09	& 40.32 & 40.32\\
      testes                & \textbf{64.20}	& 46.91	& 46.91	& 51.25 & 52.50	& 44.44	& 34.57	& 45.68	& 54.73	& 43.21 & 44.44\\
      \rowcolor{gray!15}
      uterus                & 52.31	& 40.00	& 46.15	& 46.88 & 42.19	& 41.54	& 32.31	& \textbf{53.85}	& 45.13	& 38.46 & 45.64\\
      \midrule
      \multicolumn{11}{l}{\textbf{respiratory system}} \\
      \midrule
      \rowcolor{gray!15}
      trachea bronchi       & 50.00	& 60.00	& 55.56	& 62.50 & 55.56	& 70.00	& 30.00	& 50.00	& 46.67	& \textbf{73.33} & 66.67\\
      lung                  & \textbf{59.11}	& 47.29	& 50.25	& 53.00 & 50.51	& 48.28	& 35.96	& 45.32	& 47.62	& 46.96 & 43.68\\
      \rowcolor{gray!15}
      pleura                & \textbf{52.94}	& 23.53	& 41.18	& 35.29 & 25.00	& 47.06	& 47.06	& \textbf{52.94}	& 35.29	& 37.25 & 37.25\\
      \midrule
      \multicolumn{11}{l}{\textbf{urinary system}} \\
      \midrule
      \rowcolor{gray!15}
      ureters               & 44.59	& 44.59	& 40.54	& 46.48 & 42.65	& 40.54	& 25.68	& 45.95	& 41.89	& 37.84 & \textbf{48.20}\\
      kidneys               & 50.00	& 51.19	& \textbf{58.33}	& 50.00 & 54.32	& 50.00	& 38.10	& 46.43	& 44.84	& 40.48 & 34.52\\
      \rowcolor{gray!15}
      urethra               & 52.17	& 43.48	& \textbf{60.87}	& 43.48 & 40.91	& 21.74	& 47.83	& 26.09	& 52.17	& 49.28 & 36.23\\
      bladder               & 51.43	& 57.14	& 54.29	& \textbf{65.71} & 54.29	& 51.43	& 42.86	& 40.00	& 51.43	& 36.19 & 46.67\\
      \bottomrule
    \end{tabular}
    \caption{\textbf{Accuracy of Models by organs on \OurBench.}
    We report the organ-wise accuracy of the models on \OurBench.
    The best accuracy is highlighted in bold.}
    \label{tab:comparison_of_organs}
  \vspace*{\fill}
  \end{table*}
\clearpage
\section{Prompt Details}\label{sec:prompt_details}
\begin{figure*}[p]
  \centering
  \vspace*{\fill}
  \includegraphics[width=\textwidth,height=0.8\textheight,keepaspectratio]{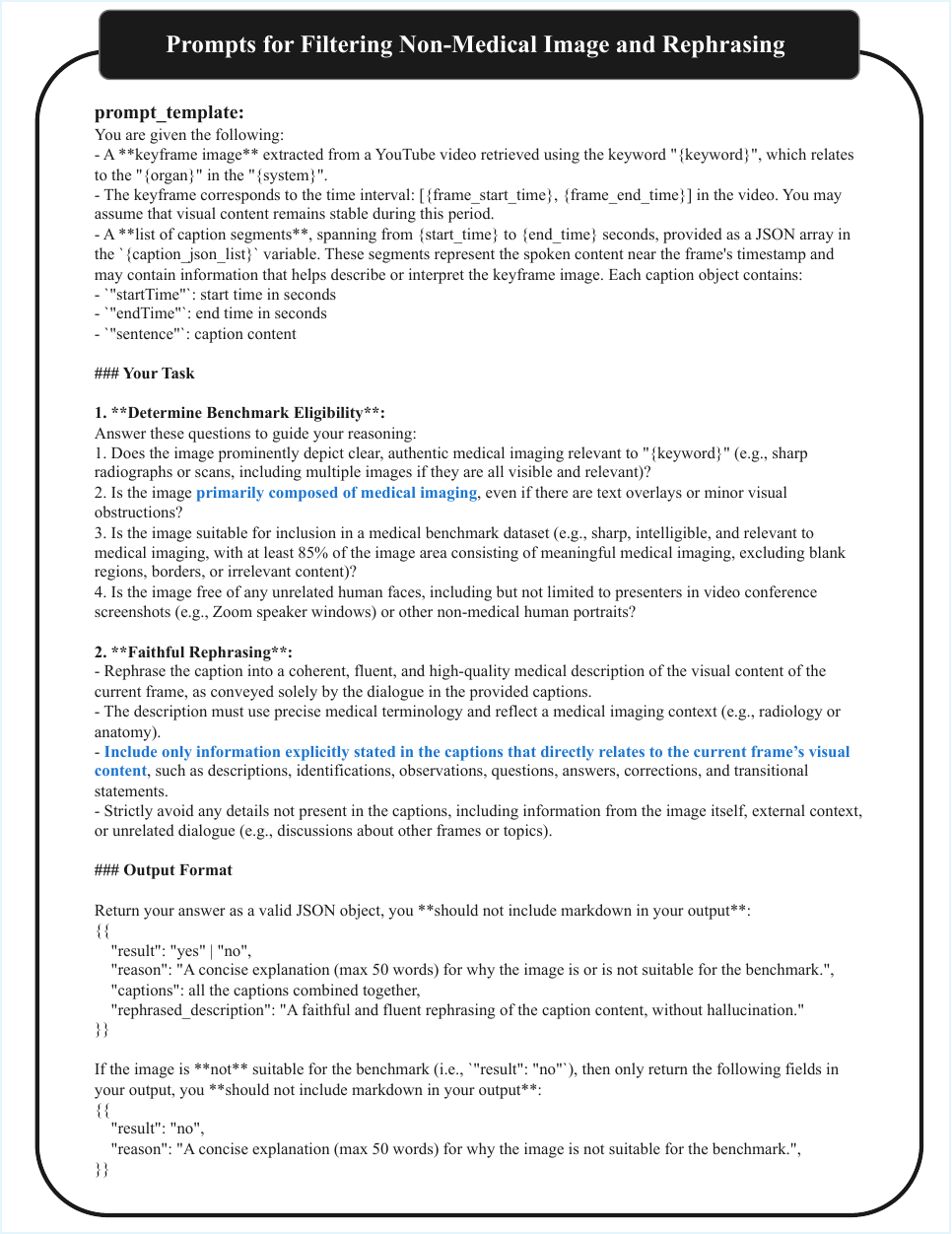}
  \par\medskip
  \caption{Filter and Rephrase Captions}
  \label{fig:filter_prompt}
  \vspace*{\fill}
\end{figure*}

\begin{figure*}[p]
  \centering
  \vspace*{\fill}
  \includegraphics[width=\textwidth,height=0.8\textheight,keepaspectratio]{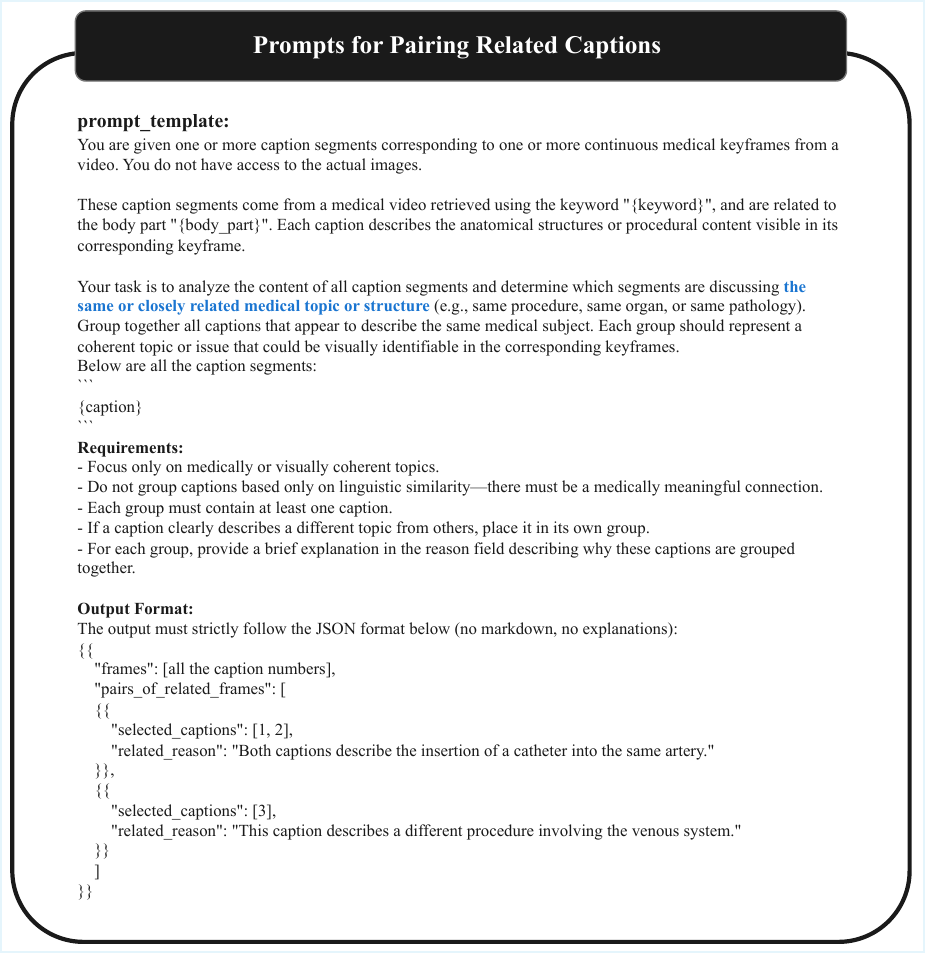}
  \caption{Transcripts Relation Check}
  \label{fig:relate_prompt}
  \vspace*{\fill}
\end{figure*}

\begin{figure*}[p]
  \centering
  \vspace*{\fill}
  \includegraphics[width=\textwidth,height=0.8\textheight,keepaspectratio]{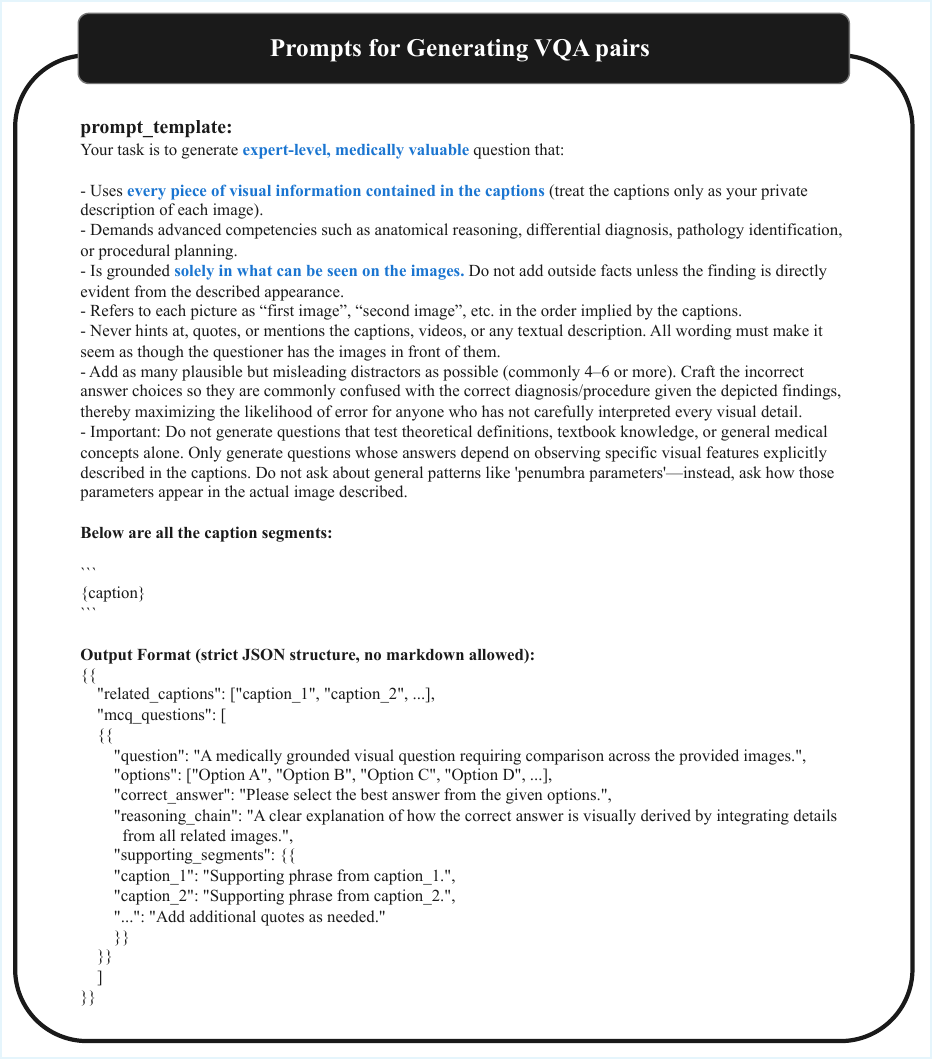}
  \caption{Multi-Frame VQA Pair Generation}
  \label{fig:generation_prompt}
  \vspace*{\fill}
\end{figure*}

\begin{figure*}[p]
  \centering
  \vspace*{\fill}
  \includegraphics[width=\textwidth,height=0.8\textheight,keepaspectratio]{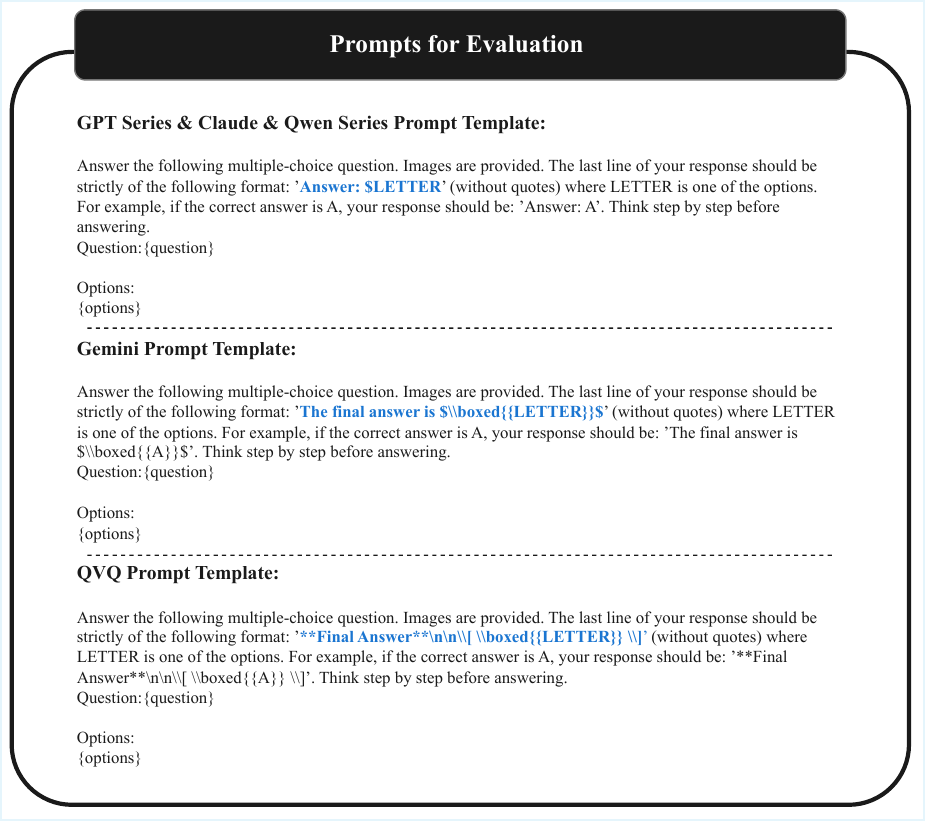}
  \caption{Benchmark Evaluation}
  \label{fig:evaluation_prompt}
  \vspace*{\fill}
\end{figure*}

We details the specific prompts and instructions utilized throughout \OurBench data curation pipeline.
We provide the full text of the prompts for frame filtering in \cref{fig:filter_prompt}, multi-frame merging in \cref{fig:relate_prompt}, and question generation in \cref{fig:generation_prompt} to facilitate reproducibility.
Additionally, the prompt template used for the model evaluation is presented in \cref{fig:evaluation_prompt}.
\clearpage

\clearpage

\clearpage

\clearpage
\section{Representative Examples}\label{sec:representative_examples}

\begin{figure*}[p]
  \centering
  \vspace*{\fill}
  \includegraphics[width=\textwidth,height=0.8\textheight,keepaspectratio]{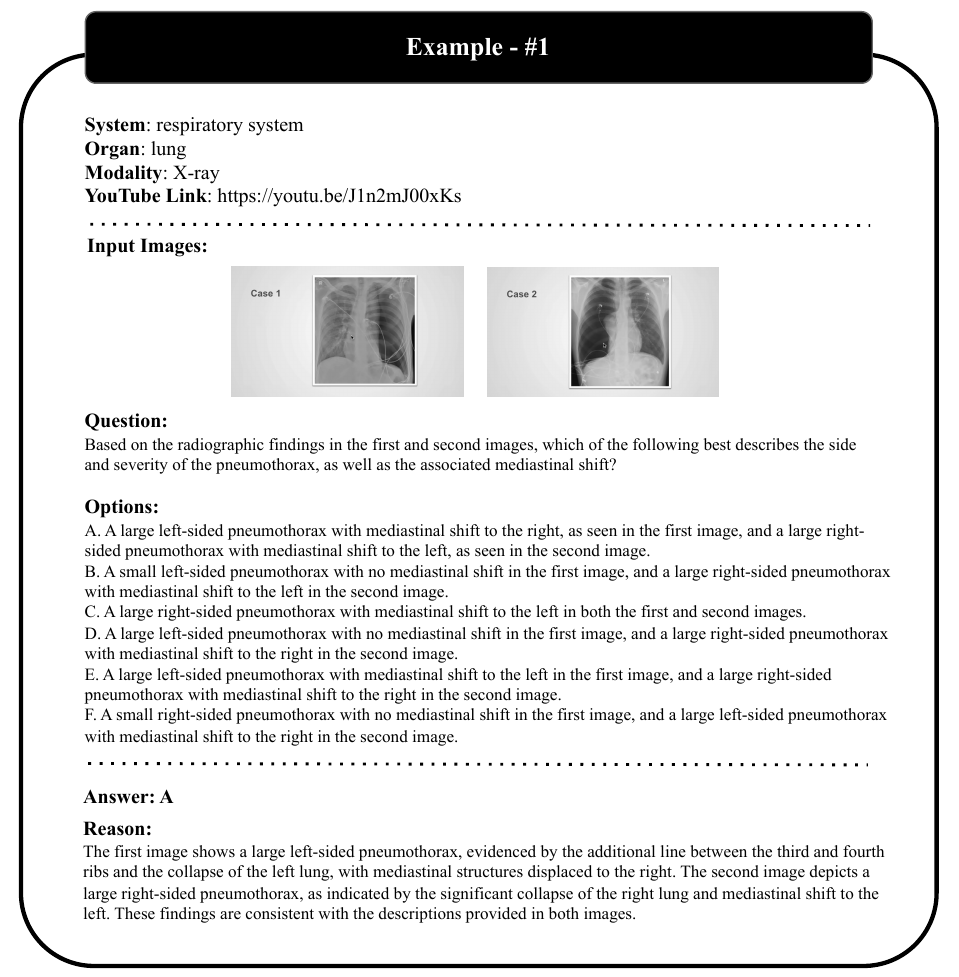}
  \par\medskip
  \caption{Two Frames Example}
  \label{fig:two_frame_example}
  \vspace*{\fill}
\end{figure*}

\begin{figure*}[p]
  \centering
  \vspace*{\fill}
  \includegraphics[width=\textwidth,height=0.8\textheight,keepaspectratio]{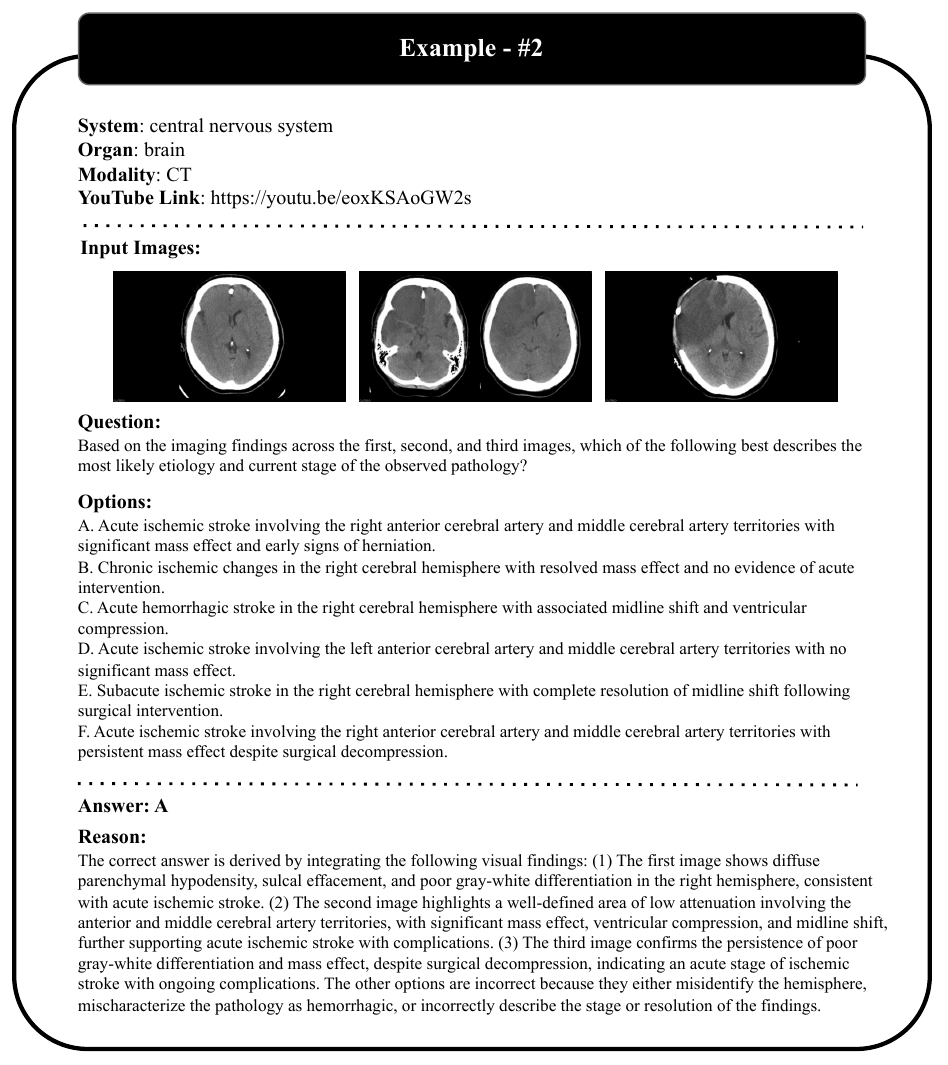}
  \caption{Three Frames Example}
  \label{fig:three_frame_example}
  \vspace*{\fill}
\end{figure*}

\begin{figure*}[p]
  \centering
  \vspace*{\fill}
  \includegraphics[width=\textwidth,height=0.8\textheight,keepaspectratio]{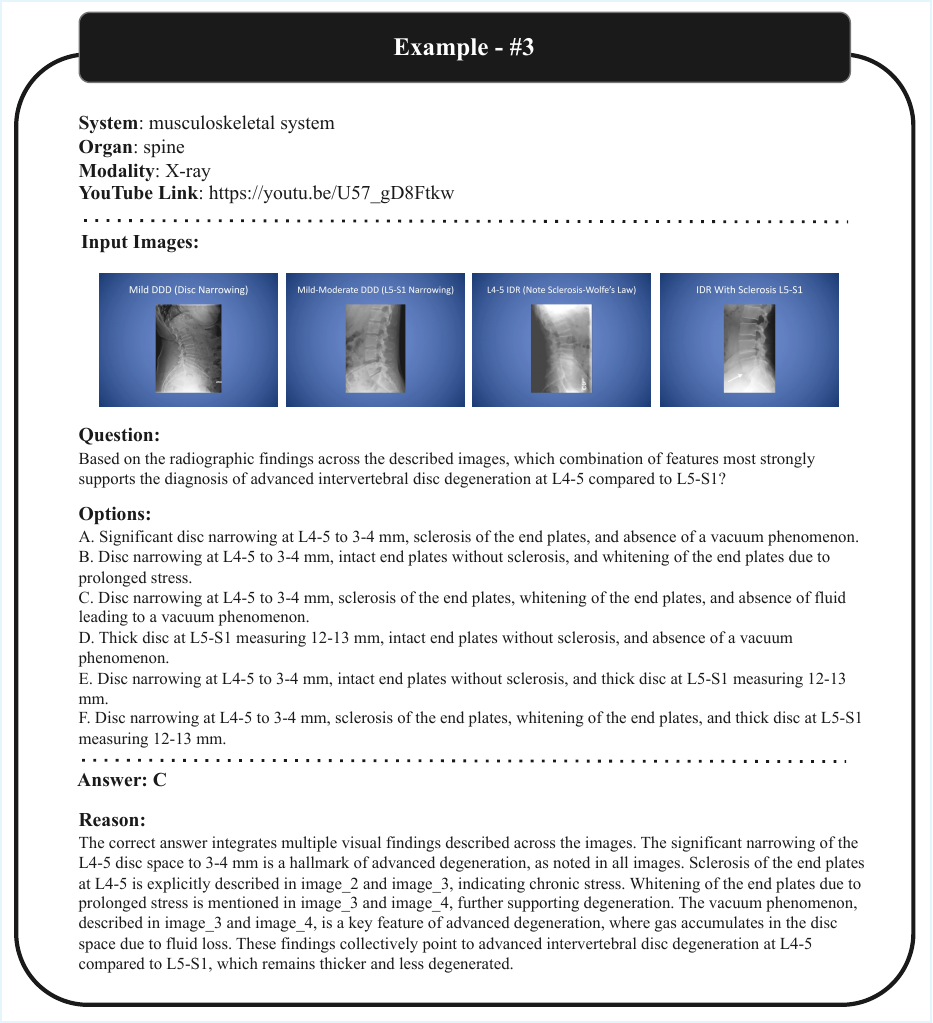}
  \caption{Four Frames Example}
  \label{fig:four_frame_example}
  \vspace*{\fill}
\end{figure*}

\begin{figure*}[p]
  \centering
  \vspace*{\fill}
  \includegraphics[width=\textwidth,height=0.8\textheight,keepaspectratio]{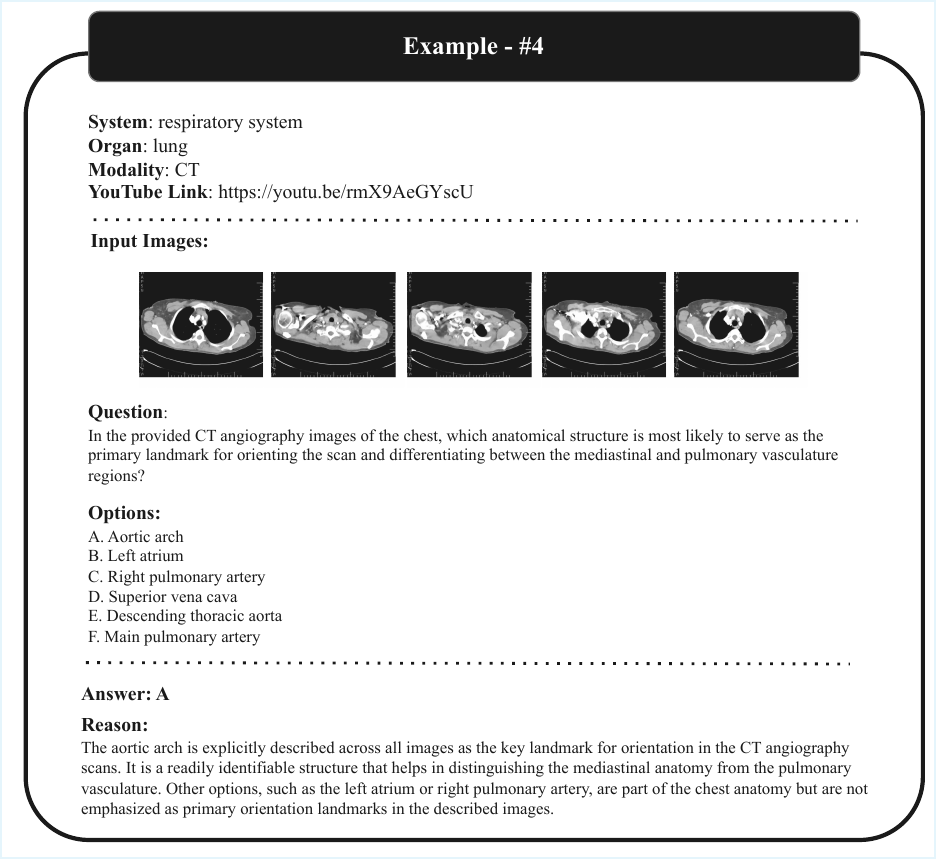}
  \caption{Five Frames Example}
  \label{fig:five_frame_example}
  \vspace*{\fill}
\end{figure*}

We provide a comprehensive visualization of representative samples from \OurBench.
Figures \cref{fig:two_frame_example} through \cref{fig:five_frame_example} showcase VQA pairs organized by varying input number of images, explicitly covering cases with two, three, four, and five frames respectively.
Each example displays the full multi-image input sequence alongside the associated clinical question and the ground-truth answer and reasoning process.
These samples highlight the complex spatial-temporal reasoning required to solve questions across different frame counts, distinguishing our benchmark from traditional single-image datasets.

\clearpage

\clearpage

\clearpage

\end{document}